\relax
\documentclass[letterpaper]{article} 

\usepackage{styleis}  
\usepackage{times}  
\usepackage{helvet} 
\usepackage{courier}  
\usepackage[hyphens]{url}  
\usepackage{graphicx} 
\usepackage{natbib}  
\usepackage{caption} 

\usepackage{amssymb}
\usepackage{amsmath}
\usepackage{amsopn}
\usepackage{epsfig}
\usepackage{microtype}
\usepackage{subfigure}
\usepackage{booktabs}
\usepackage{booktabs}
\usepackage{amsfonts}
\usepackage{nicefrac}
\usepackage{microtype}
\usepackage{algorithm}
\usepackage{rotating}
\usepackage{algorithmic}
\usepackage{wrapfig}
\usepackage{xcolor}
\usepackage{stmaryrd}
\usepackage{lipsum}
\usepackage{float}

\newfloat{algorithm}{t}{lop}

\everypar{\looseness=-1}

\newenvironment{proof}{\paragraph{Proof:}}{\hfill$\square$}
\newenvironment{subproof}{\paragraph{Proof:}}{\hfill$\blacksquare$}

\newtheorem{th_converge}{Theorem}

\newtheorem{lem_geom_arg}{Lemma}
\newtheorem{lem_analog_kir2_lemma2}[lem_geom_arg]{Lemma}

\newtheorem{def_epsequivdim}{Definition}

\newcommand{\bigo}[0]{\mathcal{O}}

\newcommand{\tsp}[0]{{\rm T}}

\newcommand{\func}[1]{{\mathfrak{#1}}}

\newcommand{\infset}[1]{{\mathbb{#1}}}
\newcommand{\finset}[1]{{\tt #1}}
\newcommand{\distrib}[1]{{\mathcal{#1}}}
\newcommand{\normdist}[0]{\distrib{N}}

\newcommand{\gp}[0]{\mathcal{GP}}
\newcommand{\rkhs}[2]{\mathcal{H}_{#1} \left( \infset{#2} \right)}

\newcommand{\sgp}[1]{\distrib{G}_{\kappa,#1}}

\DeclareMathOperator\spn{span}

\DeclareMathOperator\argmax{argmax}

\definecolor{Gray}{gray}{0.9}

\title{Sequential Subspace Search for Functional Bayesian Optimization Incorporating Experimenter Intuition}

\author{
Alistair Shilton, 
Sunil Gupta, 
Santu Rana, 
Svetha Venkatesh \\}

\affiliations{
 Applied Artificial Intelligence Institute (A2I2) \\
 Deakin University, Geelong, Australia \\
 {\tt alistair.shilton@deakin.edu.au}
}

\begin{document}

\maketitle

\begin{abstract}
We propose an algorithm for Bayesian functional optimisation - that is, finding the function to optimise a process - guided by experimenter beliefs and intuitions regarding the expected characteristics (length-scale, smoothness, cyclicity etc.) of the optimal solution encoded into the covariance function of a Gaussian Process.  Our algorithm generates a sequence of finite-dimensional random subspaces of functional space spanned by a set of draws from the experimenter's Gaussian Process.  Standard Bayesian optimisation is applied on each subspace, and the best solution found used as a starting point (origin) for the next subspace.  Using the concept of effective dimensionality, we analyse the convergence of our algorithm and provide a regret bound to show that our algorithm converges in sub-linear time provided a finite effective dimension exists.  We test our algorithm in simulated and real-world experiments, namely blind function matching, finding the optimal precipitation-strengthening function for an aluminium alloy, and learning rate schedule optimisation for deep networks.
\end{abstract}

\section{Introduction}

Functional optimisation arises in circumstances where we seek to optimise 
continuously varying phenomena.  For example we may wish to optimise the curve 
of an aeroplane's wing to minimise drag and maximise lift, define the optimal 
tempering profile (temperature as a function of time) to maximise the strength 
of an alloy, or find the activation function that works best in a neural network. 
A common characteristics in these examples is that evaluating the performance 
of a particular function is (a) expensive (for example fabricating a wing or 
training and evaluating a deep network) and (b) results in a noisy measurement.  
Furthermore we often have a beliefs regarding the characteristics that will 
perform best in a given circumstance.  For example physical intuition may tell 
us that a plane's wing should vary on a length-scale of meters, and that sharp 
points are likely to degrade performance.

Two related works in the area of functional Bayesian optimisation are 
\cite{Vie1} and control function optimisation \cite{Vel1}.  In \cite{Vie1} 
functions are represented as elements in a reproducing kernel Hilbert space 
(RKHS).  At each iteration an acquisition functional is optimised using 
(Frechet) gradient descent (with multi-start at a random initial point to avoid 
local minima), the objective evaluated, models updated and the process 
repeated.  However, experimenter beliefs about the solution only guide the 
optimisation procedure indirectly.  Alternatively, \cite{Vel1} searches the 
space of Bernstein polynomials of (at most) a particular degree, where strong 
{\em shape function} constraints can be enforced.  Shape priors include 
priors on monotonicity, unimodality, and other properties that may be expressed 
as constraints on the Bernstein basis of the solution, but not the looser 
beliefs that we are concerned with here (e.g. the expected length-scale of 
variation of a plane's wing, the smoothness of our solution (lack of sharp 
edges, or otherwise), stationarity of form etc).

In this paper we propose an algorithm to solve expensive functional 
optimisation problems with beliefs on the solution expressed as a Gaussian 
Process covariance function, allowing us to encode ``loose'' beliefs and 
intuitions regarding for example length-scales, smoothness, and cyclicity of 
the optimal solution.  We note in passing that, while our primary focus is on 
the encoding of beliefs/intuitions of this form, in principle it is possible to 
encode harder ``shape priors'' using our approach.  For example monotonicity 
may be enforced in the Gaussian Process \cite{Rii1}, and some relevant physical 
constraints may be directly built into covariance functions \cite{Jid1}.  We 
also provide a sub-linear regret bound to assure the performance of our 
algorithm through the concept of equivalent dimension.

The approach we take is to construct a sequence of low-dimensional search 
spaces by sampling the Gaussian process encoding our beliefs regarding the 
optimal solution to define a function basis (REMBO style \cite{Wan7}), and then 
use Bayesian optimisation to find the best solution (function) in this 
subspace.  This solution (function) then becomes the origin in our next 
(random) subspace (similar to LineBO \cite{Kir2}), and the process repeats 
until the experimental budget is exhausted.  By defining our search spaces 
using samples from a GP encoding our beliefs regarding the solution we give 
preference to subspaces that satisfy our expectations of the solution - for 
example if choose a GP with a long lengthscale SE covariance then our search 
subspaces will tend (on average) to span slowly varying, smooth functions, 
accelerating optimisation.

In our experiments, we begin by testing our algorithm on blind function 
matching.  This allows us to explore the performance of our algorithm under 
idealised conditions.  We then test of these results carry over into real-world 
conditions by considering two real-world problems, namely finding the optimal 
heat-treatment function for Al-Sc (aluminium-scandium) alloy to maximise its 
strength, and obtaining the optimal learning-rate schedule for training a deep 
network.

\subsection{Notation}

We use $\infset{N} = \{ 0,1,\ldots \}$, $\infset{N}_i = \{ 0,1,\ldots,i-1 \}$ 
and $\spn ( x_0, x_1, \ldots ) = \{ \sum_i \alpha_i x_i | 
\alpha_i \in \infset{R} \}$.   $|\finset{D}|$ is the number of elements in a 
finite set $\finset{D}$.  $L_2 (\infset{B})$ is the set of $L_2$-integrable 
functions $f : \infset{B} \to \infset{R}$, and $\rkhs{K}{B}$ the 
Reproducing-Kernel Hilbert Space \cite{Aro1} with reproducing kernel $K : 
\infset{B} \times \infset{B} \to \infset{R}$.  Column vectors are ${\bf a}, 
{\bf b}, \ldots$ and matrices ${\bf V}, {\bf W}, \ldots$, with elements $a_i, 
\ldots, W_{i,j}, \ldots$.  ${\bf a} \odot {\bf b}$ is the element-wise product. 
$\llbracket \cdot \rrbracket$ is the Iverson bracket \cite{Ive1} (for bool $q$, 
$\llbracket q \rrbracket = 1$ if $q$ true, $0$ otherwise).

\section{Problem Statement}

This paper is concerned with solving the problem:
\begin{equation}
 \begin{array}{l}
  \func{g}^* = \mathop{\rm argmax}\limits_{\func{g} \sim \gp (0,\kappa) : \left\| \func{g} \right\|_{L_2 (\infset{A})} \leq L_{\rm max}} f \left( \func{g} \right) \\
 \end{array}
 \label{eq:probdef}
\end{equation}
where $f : L_2 (\infset{A}) \to \infset{R}$ is an expensive (to evaluate) and 
noisy functional.  That is, we want to find the {\em function} $\func{g} : 
\infset{A} \to \infset{R}$ that produces the best results when applied in some 
process $f$ - e.g. we may wish to find the best activation function for 
a neural network or the best temperature profile to optimise the properties of 
an alloy.

By assuming $\func{g}^* \sim \gp (0,\kappa)$ is a draw from a zero-mean 
Gaussian Process characterised by the covariance function $\kappa : \infset{A} 
\times \infset{A} \to \infset{R}$, the experimenter may assert beliefs and 
intuitions regarding the properties of $\func{g}$ through the selection of the 
covariance $\kappa$.  For example, the length-scale, smoothness and periodicity 
characteristics of $\kappa$ control the length-scale, smoothness and 
periodicity characteristics of $\func{g}^*$, allowing the experimenter to 
specify how quickly the temperature may change in an annealing process, or how 
smooth the surface of a wing is.  This is in contrast to the strong ``shape 
priors'' of \cite{Vel1}, which allows strong constraints on $\mathfrak{g}$ 
such as monotonicity and unimodality to be enforced, but not looser beliefs 
on e.g. length-scale and periodicity.

\section{Background} \label{sec:basics}

\subsection{Gaussian Processes} \label{sec:gpdesc}

A Gaussian process $\gp ( \mu, K )$ is a distribution on a space of functions 
$f : \infset{X} \to \infset{R}$ with mean $\mu : \infset{X} \to \infset{R}$ and 
covariance $K : \infset{X} \times \infset{X} \to \infset{R}$ \cite{Mac3,Ras2}.  
Let $f \sim \gp (\mu,K)$ be a draw from a Gaussian process.  Then the posterior 
of $f$ given noisy observations $\finset{D} = \{(x_i, y_i) | y_i = f (x_i) + 
\epsilon_i, \epsilon_i \sim \distrib{N} (0,\sigma^2) \}$ is $f (x) | \finset{D} 
\sim \distrib{N} (\mu_\finset{D} (x), \sigma_\finset{D}^2 (x))$, where 
$\sigma_\finset{D}^2 (x) = K_\finset{D} (x,x)$, $K_\finset{D} (x,x')$ is the 
posterior covariance:
\begin{equation}
 {\!\!\!\!\!\!\small{
 \begin{array}{rl}
 \mu_\finset{D} \!\left( x \right) 
 &\!\!\!\!\!= \mu \!\left( x \right) + K \!\left( x,\finset{D} \right) \left( K \!\left( \finset{D},\finset{D} \right) + {\sigma}^2 {\bf I} \right)^{-1} \left( {\bf y} - \mu \!\left( \finset{D} \right) \right) \\
 K_\finset{D} \!\left( x,x' \right) 
 &\!\!\!\!\!= K \!\left( x,x' \right) \!-\! K \!\left( x,\finset{D} \right) \left( K \!\left( \finset{D},\finset{D} \right) \!+\! {\sigma} {\bf I} \right)^{-1} \!K \!\left( \finset{D},x' \right) 
 \end{array}
 }\!\!\!\!\!\!}
 \label{eq:gp_first}
\end{equation}
and we use the shorthand notations:
\[
 {\small{
 \begin{array}{l}
  \mu \left( \finset{D} \right) = \left[ \begin{array}{c} \mu \left( x_i \right) \end{array} \right]^{\tsp}_{i \in \infset{N}_{|\finset{D}|}}, 
  K \left( \finset{D},\finset{D} \right) = \left[ \begin{array}{c} K \left( x_i,x_j \right) \end{array} \right]_{i,j \in \infset{N}_{|\finset{D}|}} \\
  K \left( x,\finset{D} \right) = K \left( \finset{D},x \right)^{\tsp} = \left[ \begin{array}{c} K \left( x,x_i \right) \end{array} \right] \\
 \end{array}
 }}
\]
Note that this applies to functions $f : \infset{X} \to \infset{R}$ for any 
$\infset{X}$ on which a covariance $K : \infset{X} \times \infset{X} \to 
\infset{R}$ can be defined.

\subsection{Standard Bayesian Optimisation} \label{sec:standardbo}

Typically Bayesian optimisation is concerned with solving:
\begin{equation}
 \begin{array}{l}
  {\bf x}^* = \mathop{\rm argmax}\limits_{{\bf x} \in \infset{X} \subseteq \infset{R}^n} f \left( {\bf x} \right)
 \end{array}
 \label{eq:opt_prob}
\end{equation}
where $f$ is expensive to evaluate and observations are noisy.  
The aim is to 
solve (\ref{eq:opt_prob}) using the minimum evaluations of $f$.  Modelling $f 
\sim \gp (0,K)$ as a draw from a Gaussian process, Bayesian optimisation 
\cite{Jon1} is an iterative algorithm (algorithm \ref{alg:bayalg}) for solving 
(\ref{eq:opt_prob}).  At each iteration a (computationally cheap) surrogate 
acquisition function based on the GP model is optimised to select the next 
sample point, an observation is made at that point, and the GP model updated.  
The algorithm terminates either when some termination condition is satisfied 
or the budget (number of times $f$ may be evaluated) is reached.  Popular 
acquisition functions include probability of improvement \cite{Kus1}, expected 
improvement \cite{Moc1} and Gaussian process upper confidence bound (GP-UCB) 
\cite{Sri1}.  In this paper we use the GP-UCB acquisition function (of course 
others could be substituted):
\[
 \begin{array}{l}
  a_t \left( {\bf x} | \finset{D} \right) = \mu_{\finset{D}} \left( {\bf x} \right) + \sqrt{\beta_{t}} \sigma_{\finset{D}} \left( {\bf x} \right)
 \end{array}
\]
where $\beta_t$ are a sequence of constants (see \cite{Sri1,Bro2} for 
details).  In practice we find that the $\beta_t$ recommended by \cite[page 16]{Bro2} 
works well in our case without requiring many additional parameters to be 
selected.

\begin{algorithm}[t]
\begin{algorithmic}
 \INPUT Prior $K : \infset{X} \times \infset{X} \to \infset{R}$ on $f \sim \gp (0,K)$.
 \STATE Initial observations $\finset{D} = \{ ({\bf x}, y = f ({\bf x}) + \epsilon) | {\bf x} \sim \distrib{D}_{\infset{A}}$,
        $\epsilon \sim \normdist (0,\sigma^2) \}$ (for distribution $\distrib{D}_{\infset{A}}$).
 \STATE Modelling $f \sim \gp (0,K)$, proceed:
 \FOR{$t=0,1,\ldots$ until converged on $\infset{X}$}
 \STATE Solve ${\bf x} \leftarrow {\rm argmax}_{{\bf x} \in \infset{X}} \; a_t({\bf x}|\finset{D})$.
 \STATE Observe $y \leftarrow f ({\bf x}) + \epsilon$, $\epsilon \sim \normdist (0,\sigma^2)$.
 \STATE Update $\finset{D} \leftarrow \finset{D} \cup \{ ({\bf x}, y) \}$.
 \ENDFOR
 \STATE Return $({\bf x}^\star,y^\star) = {\argmax}_{({\bf x},y) \in \finset{D}} y$.
\end{algorithmic}
\caption{Standard Bayesian Optimisation.}
\label{alg:bayalg}
\end{algorithm}

\section{Method}

Recall that we are concerned with solving (\ref{eq:probdef}):
\[
 \begin{array}{l}
  \func{g}^* = \mathop{\rm argmax}\limits_{\func{g} \sim \gp (0,\kappa) : \left\| \func{g} \right\|_{L_2 (\infset{A})} \leq L_{\rm max}} f \left( \func{g} \right) \\
 \end{array}
\]
where $f : L_2 (\infset{A}) \to \infset{R}$ is an expensive (to evaluate) and 
noisy functional and $\func{g}^* \sim \gp (0,\kappa)$, where $\kappa$ 
characterises our expectations on the solution $\func{g}^*$ (e.g. the 
time-scale at which the temperature profile of the heat-treatment process 
varies, the smoothness of the plane's wing).  We model $f \sim \gp (0,K)$, and 
$K$ is our prior over the objective function $f$.

In standard Bayesian optimisation the search space is most often a 
finite-dimensional vector space $\infset{R}^d$.  For practical reasons (e.g. 
computational complexity of global optimisation of the acquisition function) 
early work concentrated on the low-dimensional case, roughly $d \lesssim 10$.  
Recently progress has been made in the high dimensional case, typically by the 
construction of either a single low-dimensional embedded subspace of the search 
space or a sequence of low-dimensional embedded subspaces on which standard 
(low-dimensional) optimisation may proceed.  For example, REMBO \cite{Wan7} 
constructs a single subspace by random embedding and applies Bayesian 
Optimisation to this subspace, while LineBO \cite{Kir2} constructs a sequence 
of $1$-dimensional subspaces (lines), optimising on each before proceeding to 
the next in a principled manner.

Functional Bayesian Optimisation represents the logical extension of 
high-dimensional Bayesian optimisation to the infinite dimensional case, where 
the discrete index $i \in \infset{N}_n$ identifying element $x_i$ of vector 
${\bf x} \in \infset{R}^n$ is supplanted by the continuous argument $a \in 
\infset{A}$ in the evaluation $\func{g}(a)$ of function $\func{g} \in 
L_2 (\infset{A})$.  However one may still apply subspace methods analogous to 
REMBO and LineBO - for example, as observed in \cite{Vie1}, random RKHS vectors 
may be used to define a basis for a subspace $\infset{T} \subset L_2 
(\infset{A})$, and optimisation may proceed on $\infset{T}$ as it has an 
(effectively) finite dimension.  Alternatively, \cite{Vel1} uses Bernstein 
polynomials to span a subspace $\infset{U} \subset L_2 (\infset{A})$, where 
optimisation may proceed as $\infset{U}$ has an (effectively) finite 
dimension.  However neither of these approaches provide a clear means of using 
our loose priors (as opposed to ``hard'' shape priors \cite{Vel1}) on 
$\func{g}^*$ to accelerate the optimisation procedure.

Motivated by this, our algorithm (section \ref{sec:ouralg}) is a hybrid 
extension of REMBO and LineBO.  The {\em outer loop} selects a sequence of $S$ 
($S$ is the outer-loop budget) $d$-dimensional subspaces (as in LineBO, but 
multi-dimensional) by sampling from $\gp(0,\kappa)$ to generate a finite basis 
for $\infset{U}_s = \func{b}_s + \spn (\func{h}_s^0, \func{h}_s^1, \ldots 
\func{h}_s^{d-1}) \subset L_2 (\infset{A})$, where $s \in \infset{N}_S$ is an 
iteration count, $\func{b}_s$ is the best solution found up to iteration $s$, 
and $\func{h}_s^{0}, \func{h}_s^{1}, \ldots, \func{h}_s^{d-1} \sim \gp 
(0,\kappa)$) that favours functions with the characteristics we expect in 
$\func{g}^*$, while the {\em inner loop} searches $\infset{U}_s$ using standard 
Bayesian Optimisation.  Note that:
\begin{enumerate}
 \item The algorithm uses two distinct covariance functions:
 \begin{enumerate}
 \item Covariance $\kappa$ guides subspace selection for each outer loop 
       iteration $s$, guiding the algorithm to explore subspaces of functions 
       with characteristics we expect of $\func{g}^*$.
 \item Covariance $K$ characterises the functional space, which we discuss in 
       detail in section \ref{sec:covfunc}.
 \end{enumerate}
 \item Each function $\func{g}$ evaluated in the inner loop is a 
       weighted sum of the basis functions $\func{h}_s^{0}, \func{h}_s^{1}, \ldots, 
       \func{h}_s^{d-1}$, and, recursively through the bias $\func{b}_s$, 
       all previous such bases.  This sum contains at most $dS$ terms.  In our 
       implementation we use pre-sampling and caching to avoid computational 
       issues arising from this as described in the supplementary.
 \item Convergence of the inner loop can be assessed using either a simple 
       budget of $T$ evaluations (resulting in $ST$ 
       evaluations overall over $S$ outer loop iterations) or a simple regret test 
       as per LineBO \cite{Kir2} to terminate the inner loop if ${\rm err} 
       (\func{g}_s^\star) < \epsilon$, where:
       \[
        \!\!\!\!\begin{array}{l}
         {\rm err} \left( \func{g} \right) = \mu_{\finset{D}} \left( \func{g} \right) + \sigma_{\finset{D}} \left( \func{g} \right) - {\min}_{\func{g}' \in \infset{U}_s} \left( \mu_{\finset{D}} \left( \func{g}' \right) - \sigma_{\finset{D}} \left( \func{g}' \right) \right)
        \end{array}
       \]
\end{enumerate}

\subsection{Modelling the Objective} \label{sec:covfunc}

In our algorithm we model $f$ as a draw from a zero-mean Gaussian Process $f 
\sim \gp (0,K)$, where $K : L_2 (\infset{A}) \times L_2 (\infset{A}) \to 
\infset{R}$.  This necessitates the construction of an appropriate covariance 
$K$.  Two potential approaches to constructing this covariance are:
\begin{enumerate}
 \setlength\itemsep{0em}
 \setlength\parskip{1pt}
  \item As per \cite{Vie1}, build $K : \rkhs{\kappa}{A} \times \rkhs{\kappa}{A} \to 
       \infset{R}$ on the RKHS $\rkhs{\kappa}{A}$ by taking a 
       stationary covariance on $\infset{R}^d$ and replacing $\| {\bf x} - {\bf 
       x}' \|_2^2$ with $\| \func{g} - \func{g}' \|_{\rkhs{\kappa}{A}}^2$.  For 
       example:
       \begin{equation}
        \begin{array}{l}
         K \left( \func{g},\func{g}' \right) = \exp ( -\frac{1}{2\gamma^2} \left\| \func{g} - \func{g}' \right\|_{\rkhs{\kappa}{A}}^2 ) \\
        \end{array}
        \label{eq:functional_L2}
       \end{equation}
 \item Noting that $L_2 (\infset{A})$ is a Hilbert space, build $K : L_2 
       (\infset{A}) \times L_2 (\infset{A}) \to \infset{R}$ by taking a 
       stationary covariance on $R^d$ and replacing $\| {\bf x} - {\bf x}' 
       \|_2^2$ with $\| \func{g}-\func{g}' \|_{L_2 (\infset{A})}^2$.  For 
       example:
       \begin{equation}
        \begin{array}{l}
         K \left( \func{g},\func{g}' \right) = \exp ( -\frac{1}{2\gamma^2} \left\| \func{g} - \func{g}' \right\|_{L_2 (\infset{A})}^2 )
        \end{array}
        \label{eq:functional_SE}
       \end{equation}
\end{enumerate}
Both approaches require numerical approximation.  In the first approach, 
functions $\func{g}, \func{g}' \in L_2 (\infset{A})$ must be approximated as 
$\func{g} \approx \sum_i \alpha_i \kappa (\cdot, {\bf c}^i), \func{g}' 
\approx \sum_i \alpha'_i \kappa (\cdot, {\bf c}^i)$ for a suitable grid of 
points ${\bf c}^i \in \infset{A}$ (e.g. an even grid of $N^{1/m}$ points per 
axis in $\infset{A} \subset \infset{R}^m$ with spacing $\tau$), so:
\begin{equation}
 \begin{array}{l}
  \| \func{g} - \func{g}' \|^2_{\rkhs{\kappa}{A}} \approx 
  {\sum}_{i,j} ( \alpha_i - \alpha'_i ) ( \alpha_j - \alpha'_j ) \kappa \left( {\bf c}^i, {\bf c}^j \right) \\
 \end{array}
 \label{eq:define_g_ghs}
\end{equation}
Likewise in approach 2, using the same grid, we may use a histogram 
approximation:
\begin{equation}
 \begin{array}{rl}
  \| \func{g}-\func{g}' \|_{L_2 (\infset{A})}^2 
  &\!\!\!\approx \sum_i ( \func{g} ({\bf c}^i) - \func{g}'({\bf c}^i) )^2 \tau^m
 \end{array}
 \label{eq:diffapprox}
\end{equation}

The computational cost of the first approach scales quadratically with the 
size $N$ of the grid, whereas approach $2$ scales linearly.  Furthermore 
approach 1 conflates two distinct beliefs, namely (a) our beliefs 
regarding the expected properties of the $\func{g}^* \sim \gp 
(0,\kappa)$ (smoothness, length-scale etc), and (b) our prior regarding the 
characteristics of the objective $f \sim \gp (0,K)$, as the recipe for approach 
1 embeds the former into the latter via (\ref{eq:define_g_ghs}) in the 
construction of $K$.  However there is no a-priori reason to link these 
concepts, or presume that such linking will improve convergence.

By contrast the approach we have selected (approach 2) builds $K$ from the (solution, or $\kappa$-) belief 
agnostic function-difference measure $\| \func{g}-\func{g}' \|_{L_2 
(\infset{A})}$, so $\kappa$ and $K$ serve two distinct purposes: $\kappa$ 
guides our choice of search sub-space, giving preference to subspaces containing 
mostly functions that we expect to be similar to the optimal solution 
$\func{g}^*$; and $K$ is used to model our objective function $f$.

We discuss how (\ref{eq:diffapprox}) may be efficiently computed using a grid 
approximation in the supplementary material.  For practical purposes, we note that the 
computational cost of this approximation on our algorithm scales linearly with 
the grid-size $N$, so the penalty for ``overdoing it'' to ensure an accurate 
approximation is relatively small (for example in 
our experiments we use $N = 100$ without difficulty), so we recommend being 
generous in this respect.  If the grid is too small then the effect will be 
similar to choosing a length-scale on $K$ that is too large, as an overly 
coarse grid will be unable to capture fine (sharp) features in $\func{g}$ 
(effectively calculating the difference smoothed approximations).  Finally, 
we implicitly assume a low-dimensional domain $\infset{A} = \infset{R}^\nu$ 
for $\func{g}$ (practically $\nu \leq 3$).  This captures many physical cases 
of interest like scheduling ($\nu = 1$) or wing design ($\nu = 2$) while still 
retaining a practical grid size $N = \rho^\nu$.  The extension to higher 
dimensions will require some additional approximation of $\func{g}$ to keep 
the computational cost within sensible bounds, but this is beyond the scope of 
the present paper.

\subsection{The Algorithm} \label{sec:ouralg}

Our proposed algorithm is shown in algorithm \ref{alg:ouralg}.  As 
noted previously, the {\em outer loop} selects a sequence of subspaces by 
drawing a basis $\func{h}^0_s, \func{h}^1_s, \ldots, \func{h}^{d-1}_s \sim 
\gp (0,\kappa)$ to define a subspace $\infset{U}_s = \func{b}_s + \spn (\func{h}^0_s, 
\func{h}^1_s, \ldots, \func{h}^{d-1}_s)$ that is biased to favour 
functions with the characteristics we expect of the optima $\func{g}^* \sim 
\gp (0,\kappa)$.  The {\em inner loop} uses standard Bayesian Optimisation with 
a GP-UCB acquisition function to find the best solution on this 
subspace, which then becomes the bias $\func{b}_{s+1}$ for the next outer-loop 
iteration, and so on.  The selection of the subspace dimension $d$ is discussed 
in section \ref{sec:outer_loop}, but loosely speaking $1 \leq d \leq \max 
(10,d_e)$, where $d = 1$ makes the algorithm behave like LineBO, $d = d_e$ 
makes it behave like REMBO, $d_e$ is the effective dimension of the objective 
$f$ (definition \ref{def:epsequivdim}, though this is rarely known), and $10$ 
is the practical upper-bound for computational reasons.  The hyperparameters 
for covariance $K$ are tuned for max-log-likelihood in the usual manner.

\begin{algorithm}
\caption{Sequential-Subspace-Search Bayesian Functional Optimisation 
         ($S^3$-BFO) Algorithm.\protect\footnotemark}
\label{alg:ouralg}
\begin{algorithmic}
 \INPUT Prior $K : L_{2} ({\infset{A}}) \times L_{2} ({\infset{A}}) \to \infset{R}$ on $f \sim \gp(0,K)$.
 \INPUT Prior $\kappa : \infset{A} \times \infset{A} \to \infset{R}$ on $\func{g}^* \sim \gp(0,\kappa)$.
 \STATE Let $(\func{g}_{[0]}^\star,y_{[0]}^\star) = (0,0)$.\vspace{-0.05cm}
 \STATE Modelling $f \sim \gp (0,K)$, proceed:\vspace{-0.1cm}
 \FOR{$s=0,1,\ldots,S-1\;\;\;\;\;\;\mbox{\colorbox{blue!30}{(outer loop)}}\!\!\!\!\!\!\!\!\!\!\!\!\!\!\!\!\!\!\!\!\!\!\!\!\!\!\!\!\!\!\!\!\!\!\!\!\!\!\!\!\!\!$}
 \STATE Assign $\func{b}_s \leftarrow \func{g}_{[s]}^\star$.
 \STATE Sample $\func{h}_s^{0},\func{h}_s^{1},\ldots,\func{h}_s^{d-1} \sim \gp (0,\kappa)$.
 \STATE Initial observations $\finset{D}_s = \{ (\func{g}=\func{b}_s+\sum_j \lambda_j \func{h}_s^{j}, y = f (\func{g}) + \epsilon) | {\mbox{\boldmath $\lambda$}} \sim \distrib{D}_{\infset{R}^d}$,
        $\epsilon \sim \normdist (0,\sigma^2) \}$.
 \FOR{$t=0,1,..$ until converged on $\infset{U}_s\;\;\;\;\;\;\mbox{\colorbox{green!30}{(inner loop)}}\!\!\!\!\!\!\!\!\!\!\!\!\!\!\!\!\!\!\!\!\!\!\!\!\!\!\!\!\!\!\!\!\!\!\!\!\!\!\!\!\!\!$}
 \STATE Solve ${\mbox{\boldmath $\lambda$}} \leftarrow {\rm argmax}_{{{\mbox{\boldmath\scriptsize $\lambda$}} \in \infset{R}^d}} a_t(\func{b}_s+\sum_j \lambda_j \func{h}_s^{j}|\finset{D}_{[s]} \cup \finset{D}_s)$.
 \STATE Project $\func{g} \leftarrow \func{b}_s+\sum_j \lambda_j \func{h}_s^{j}$.
 \STATE Observe $y \leftarrow f (\func{g}) + \epsilon$, $\epsilon \sim \normdist (0,\sigma^2)$.
 \STATE Update $\finset{D}_s \leftarrow \finset{D}_s \cup \{ (\func{g},y) \}$.
 \ENDFOR
 \STATE Let $\finset{D}_{[s+1]} = \finset{D}_{[s]} \cup \finset{D}_{s}$.
 \STATE Let $(\func{g}_{[s+1]}^\star,y_{[s+1]}^\star) = {\argmax}_{(\func{g},y) \in \finset{D}_{[s+1]}} y$.
 \ENDFOR
 \STATE Return $(\func{g}_{[S]}^\star,y_{[S]}^\star)$.
\end{algorithmic}
\end{algorithm}
\footnotetext{For clarity, when reading this algorithm, note that the 
super/subscript $s$ implies ``for iteration $s$'', whereas the super/subscript 
$[s]$ implies ``up to but not including iteration $s$'', so for example 
$\finset{D}_s$ is the set of observations made during iteration $s$, whereas 
$\finset{D}_{[s]}$ is the set of all observations made prior to iteration $s$.}

\section{Convergence Analysis}

As in the analysis of REMBO and LineBO, our convergence analysis is based 
around the concept of {\em effective dimensionality}.  For the functional case 
we define this as follows:
\begin{def_epsequivdim}
 Let $f : L_2 (\infset{A}) \to \infset{R}$.  The effective dimension of $f$ is 
 the lowest $d_e \in \infset{N}$ such that there exists $\bar{\func{h}}^0, 
 \bar{\func{h}}^1, \ldots, \bar{\func{h}}^{d_e-1} \in L_2 (\infset{A})$ 
 such that $\| f (\func{g} + \func{g}_{\perp}) - f (\func{g}) \|_{L_2 
 (\infset{A})}  = 0$ $\forall \func{g} \in \infset{T}$, $\forall 
 \func{g}_{\perp} \in \infset{T}^{\perp}$, where $\infset{T} = \spn 
 (\bar{\func{h}}^0, \bar{\func{h}}^1, \ldots, \bar{\func{h}}^{d_e-1})$.
 \label{def:epsequivdim}
\end{def_epsequivdim}

In our analysis of algorithm \ref{alg:ouralg} we first consider the inner and 
outer loops separately.  The inner-loop may be analysed in terms of the 
standard BO optimisation using GP-UCB acquisition function \cite{Sri1}; whereas 
the outer loop analysis more closely models the analysis of LineBO in 
\cite{Kir2}.

\subsection{Inner-Loop Convergence}

Our aim here is to bound cumulative regret bound $R_t = \sum_t ( 
f(\func{g}^*) - f(\func{g}^t) )$ on the inner loop of algorithm 
\ref{alg:ouralg} in terms of the inner-loop iteration counter $t$.  The 
complicating factors are:
\begin{enumerate}
 \setlength\itemsep{0em}
 \setlength\parskip{1pt}
  \item The model is not built on the variables optimised by the inner loop (the 
       ${\mbox{\boldmath $\lambda$}}$'s) but rather the projection of these 
       objects into function space (the $\func{g}$'s).
 \item The model used for $f$ is built from not just the current instance of 
       the inner loop but all previous instances.
\end{enumerate}

With regard to point 1, note that, in terms of our basis $\infset{U}_s$, we can 
rewrite $\| \func{g} - \func{g}' \|_2^2$ as:
\[
 {\small{
 \begin{array}{rll}
  \left\| \func{g} - \func{g}' \right\|_{L_2 (\infset{A})}^2 
  &\!\!\!\!= \left\| \sum_i \left( \lambda_i - \lambda'_i \right) \func{h}^i_s \right\|_{L_2 (\infset{A})}^2  \\
  &\!\!\!\!= \left( {\mbox{\boldmath $\lambda$}} - {\mbox{\boldmath $\lambda$}}' \right)^\tsp {\bf H}^s \left( {\mbox{\boldmath $\lambda$}} - {\mbox{\boldmath $\lambda$}}' \right) \\
 \end{array}
 }}
\]
where ${\bf H}^s \succeq {\bf 0}$, $H_{ij}^s = \left< \func{h}^i_s,\func{h}^j_s 
\right>_{L_2 (\infset{A})}$, which has the form of a Mahalanobis distance.  But 
this is equivalent to a standard Euclidean distance operating on data that has 
been appropriately scaled and rotated.  In particular, the maximum information 
gain \cite{Cov2} $\gamma_t$ of a covariance function depends only on the number 
of observations and not how they have been rotated and/or scaled.  Thus if we 
construct our covariance function $K$ by taking a translation-invariant 
covariance on $\infset{R}^d$ with now maximum information gain and then 
translating it to a covariance on $L_2 (\infset{A})$ then the maximum 
information gain of the resulting covariance will be the same as for the 
original.

\begin{figure*}
\centering
\[
 \begin{array}{l}
 \mbox{By definition, the posterior mean and variance of $f \sim \gp (0,K)$ given $\finset{D}_{[s]} \cup \finset{D}_s$ are: } \\
 {\small{
 \begin{array}{rl}
 \mu_{\finset{D}_{[s]} \cup \finset{D}_s} \left( \func{g} \right) 
 &\!\!\!= \left[ \begin{array}{c} K \left( \finset{D}_{[s]},\func{g} \right) \\ K \left( \finset{D}_s,\func{g} \right) \\ \end{array} \right]^{\tsp} \left[ \begin{array}{cc} K \left( \finset{D}_{[s]}, \finset{D}_{[s]} \right) + {\sigma}^2 {\bf I} & K \left( \finset{D}_{[s]}, \finset{D}_s \right) \\ K \left( \finset{D}_s, \finset{D}_{[s]} \right) & K \left( \finset{D}_s, \finset{D}_s \right) + {\sigma}^2 {\bf I} \\ \end{array} \right]^{-1} \left[ \begin{array}{c} {\bf y}_{\finset{D}_{[s]}} \\ {\bf y}_{\finset{D}_s} \\ \end{array} \right] \\
 \sigma_{\finset{D}_{[s]} \cup \finset{D}_s}^2 \left( \func{g} \right) 
 &\!\!\!= K \left( \func{g},\func{g} \right) - \left[ \!\!\begin{array}{c} K \left( \finset{D}_{[s]},\func{g} \right) \\ K \left( \finset{D}_s,\func{g} \right) \\ \end{array} \!\!\right]^{\tsp} \left[ \!\!\begin{array}{cc} K \left( \finset{D}_{[s]}, \finset{D}_{[s]} \right) \!+\! {\sigma}^2 {\bf I} & K \left( \finset{D}_{[s]}, \finset{D}_s \right) \\ K \left( \finset{D}_s, \finset{D}_{[s]} \right) & K \left( \finset{D}_s, \finset{D}_s \right) \!+\! {\sigma}^2 {\bf I} \\ \end{array} \!\!\right]^{-1} \left[ \!\!\begin{array}{c} K \left( \finset{D}_{[s]},\func{g} \right) \\ K \left( \finset{D}_s,\func{g} \right) \\ \end{array} \!\!\right] \\
 \end{array} 
 }} \\
 \mbox{Likewise, the posterior mean and covariance of $f \sim \gp (0,K)$ given only $\finset{D}_{[s]}$ are: }^{{\;}^{{{\;}^{{{\;}^{{{\;}^{\;}}}}}}}} \\
 {\small{
 \begin{array}{rl}
 \mu_{\finset{D}_{[s]}} \left( \func{g} \right) 
 &\!\!\!= K \left( \func{g},\finset{D}_{[s]} \right) \left( K \left( \finset{D}_{[s]}, \finset{D}_{[s]} \right) + {\sigma}^2 {\bf I} \right)^{-1} {\bf y}_{\finset{D}_{[s]}} \\
 K_{\finset{D}_{[s]}}^2 \left( \func{g}, \func{g}' \right) 
 &\!\!\!= K \left( \func{g},\func{g}' \right) -  K \left( \func{g},\finset{D}_{[s]} \right) \left( K \left( \finset{D}_{[s]}, \finset{D}_{[s]} \right) + {\sigma}^2 {\bf I} \right)^{-1} K \left( \finset{D}_{[s]},\func{g}' \right) \\
 \end{array} 
 }} \\
 \mbox{Using the matrix inversion lemma, it is straightforward to rewrite the former in terms of the latter: }^{{\;}^{{{\;}^{{{\;}^{{{\;}^{\;}}}}}}}} \\
 {\small{
 \begin{array}{rl}
 \mu_{\finset{D}_{[s]} \cup \finset{D}_s} \left( \func{g} \right) 
 &\!\!\!= \mu_{\finset{D}_{[s]}} \left( \finset{D}_s \right) + K_{\finset{D}_{[s]}} \left( \func{g},\finset{D}_s \right) \left( K_{\finset{D}_{[s]}} \left( \finset{D}_s, \finset{D}_s \right) + {\sigma}^2 {\bf I} \right)^{-1} \left( {\bf y}_{\finset{D}_s} - \mu_{\finset{D}_{[s]}} \left( \finset{D}_{s} \right) \right) \\
 \sigma_{\finset{D}_{[s]} \cup \finset{D}_s}^2 \left( \func{g} \right) 
 &\!\!\!= K_{\finset{D}_{[s]}} \left( \func{g},\func{g} \right) - K_{\finset{D}_{[s]}} \left( \func{g},\finset{D}_s \right) \left( K_{\finset{D}_{[s]}} \left( \finset{D}_s, \finset{D}_s \right) + {\sigma}^2 {\bf I} \right)^{-1} K_{\finset{D}_{[s]}} \left( \finset{D}_s,\func{g} \right) \\
 \end{array} 
 }} \\
 \end{array}
\]
\caption{Derivation of posterior distribution of $f$ in the inner loop of algorithm \ref{alg:ouralg} in terms of $\finset{D}_s$.}
\label{fig:big_eqn}
\end{figure*}

With regard to point 2, recall that the posterior of $f \sim \gp (0,K)$ at 
outer iteration $s$ prior to entering the inner loop is $f(\func{g}) | 
\finset{D}_{[s]} \sim \distrib{N} (\mu_{\finset{D}_{[s]}} (\func{g}), 
\sigma^2_{\finset{D}_{[s]}} (\func{g}))$ as per (\ref{eq:gp_first}).  
Similarly, at iteration $t$ in the inner loop the posterior of $f \sim \gp 
(0,K)$ is $f(\func{g}) |\finset{D}_{[s]} \cup \finset{D}_s \sim \distrib{N} 
(\mu_{\finset{D}_{[s]} \cup \finset{D}_s} (\func{g}),\sigma^2_{\finset{D}_{[s]} 
\cup \finset{D}_s} (\func{g}))$ as per (\ref{eq:gp_first}).  
See figure \ref{fig:big_eqn}, it is not difficult to show that this posterior is 
equivalent to the posterior of $f \sim \gp(\mu_{\finset{D}_{[s]}}, 
K_{\finset{D}_{[s]}})$ given $\finset{D}_s$ - the posterior of the 
biased GP whose prior covariance is $K_{\finset{D}_{[s]}}$.

Hence, for outer-loop iteration $s$, the inner loop is essentially standard 
GP-UCB BO, \cite{Sri1}, but with covariance prior $K_{\finset{D}_{[s]}}$.  
Denoting by $\gamma_{\finset{D}_{[s]},t}$ the maximum information gain for this 
covariance function, we have from \cite{Sri1}, theorem 2, that the regret for 
the inner loop of our algorithm goes as:
\[
 \begin{array}{l}
  \bigo^* \left( \sqrt{dt\gamma_{\finset{D}_{[s]},t}} \right)
 \end{array}
\]
where we follow the notation of 
\cite{Sri1} in using $\bigo^*$ to denote $\bigo$ with log factors suppressed.  
Let $\gamma_t$ be the maximum information gain for covariance $K$.  Clearly 
$\gamma_{\finset{D}_{[s]},t} = \gamma_{sT+t}$, and moreover we have seen that 
the maximum information gain of $K$ is precisely the maximum information gain 
of the non-functional covariance from which it was constructed.  Hence for some 
standard covariance functions we have the bounds \cite{See1,Sri1}:
\begin{itemize}
 \setlength\itemsep{0em}
 \setlength\parskip{1pt}
 \item Linear: $\gamma_{\finset{D}_{[s]},t} \in \bigo (d\log t)$.
 \item Squared exponential: $\gamma_{\finset{D}_{[s]},t} \in \bigo ((\log t)^{d+1})$.\vspace{-0.1cm}
 \item Matern $\nu > 1$: $\gamma_{\finset{D}_{[s]},t} \in \bigo (t^{\frac{d(d+1)}{2\nu+d(d+1)}}\log t)$.
\end{itemize}
\vspace{-0.2cm}
where we have used the fact that $\log (sT+t) = \log t + \log (1+\frac{sT}{t}) 
\in \bigo (\log t)$ in this construction ($s$ being fixed for any given instance 
of the inner loop).

\subsection{Outer-Loop and Overall Convergence} \label{sec:outer_loop}

With regard to the outer-loop convergence we have the following result, which 
is analogous to Proposition 1 in \cite{Kir2} and considers on the number of 
(outer-loop) iterations the algorithm performs:
\begin{th_converge}
 Let $f \sim \gp (0,K)$ be a draw from a Gaussian Process with twice 
 Frechet-differentiable covariance $K : L_2(\infset{A}) \times L_2(\infset{A}) 
 \to \infset{R}$ with effective dimension $d_e$ and maxima $\func{g}^* = 
 \arg\max_{\func{g} \in L_2 (\infset{A})} f(\func{g})$, and let $\delta \in 
 (0,1)$.  Then, using the notation of algorithm \ref{alg:ouralg}, after $s$ 
 (outer-loop) iterations of algorithm \ref{alg:ouralg}, with probability at 
 least $1-\delta$:
 \[
  \begin{array}{l}
   f \left( \func{g}^* \right) - f \big( \func{g}_{[s]}^\star \big) \in \bigo \big( \llbracket d < d_e \rrbracket \left( \frac{1}{s} \log \left( \frac{1}{\delta} \right) \right)^{\frac{2}{d_e-d}} + \epsilon_{d,\delta} \big)
  \end{array}
 \]
 where $\epsilon_{d,\delta}$ is the (order-of) regret bound for the inner-loop 
 (on the subspace $\infset{U}_s = \func{b}_s + \spn(\func{h}_s^0, \func{h}_s^1, 
 \ldots, \func{h}_s^{d-1})$, not the whole space $L_2 (\infset{A})$), 
 $(\func{g}_{[s]}^\star, y_{[s]}^\star)$ is the best solution found up to the 
 start of iteration $s$, and $\llbracket \cdot \rrbracket$ is the Iverson 
 bracket.
 \label{thm:th_converge}
\end{th_converge}
\begin{proof}
The proof may be found in the supplementary material.  It is based around the 
proof of proposition 1 from \cite{Kir2}, with some novel aspects.
\end{proof}

\begin{figure*}
 \centering 
 \begin{turn}{90}
 $\;\;\;\;\;\;\;\;\;\;\;\;\;$ {\small $\scriptscriptstyle{\| \func{g}-\func{q} \|_{L_2 ([0,1])}}$}
 \end{turn}
  \includegraphics[width=0.27\textwidth]{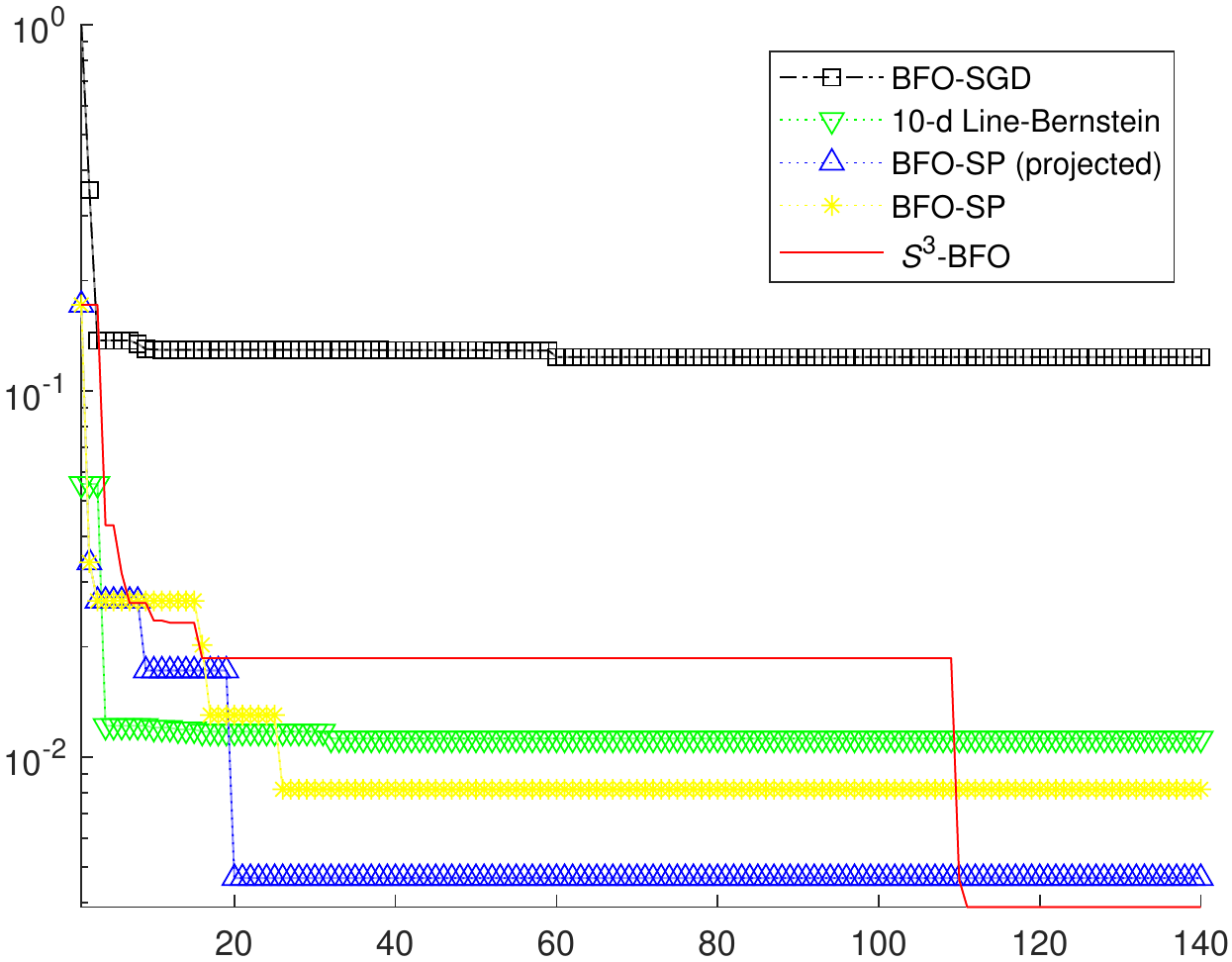}   
 $\;$ \begin{turn}{90}
 $\;\;\;\;\;\;\;\;\;\;\;\;\;$ {\small $\scriptscriptstyle{\| \func{g}-\func{q} \|_{L_2 ([0,1])}}$}
 \end{turn}
  \includegraphics[width=0.27\textwidth]{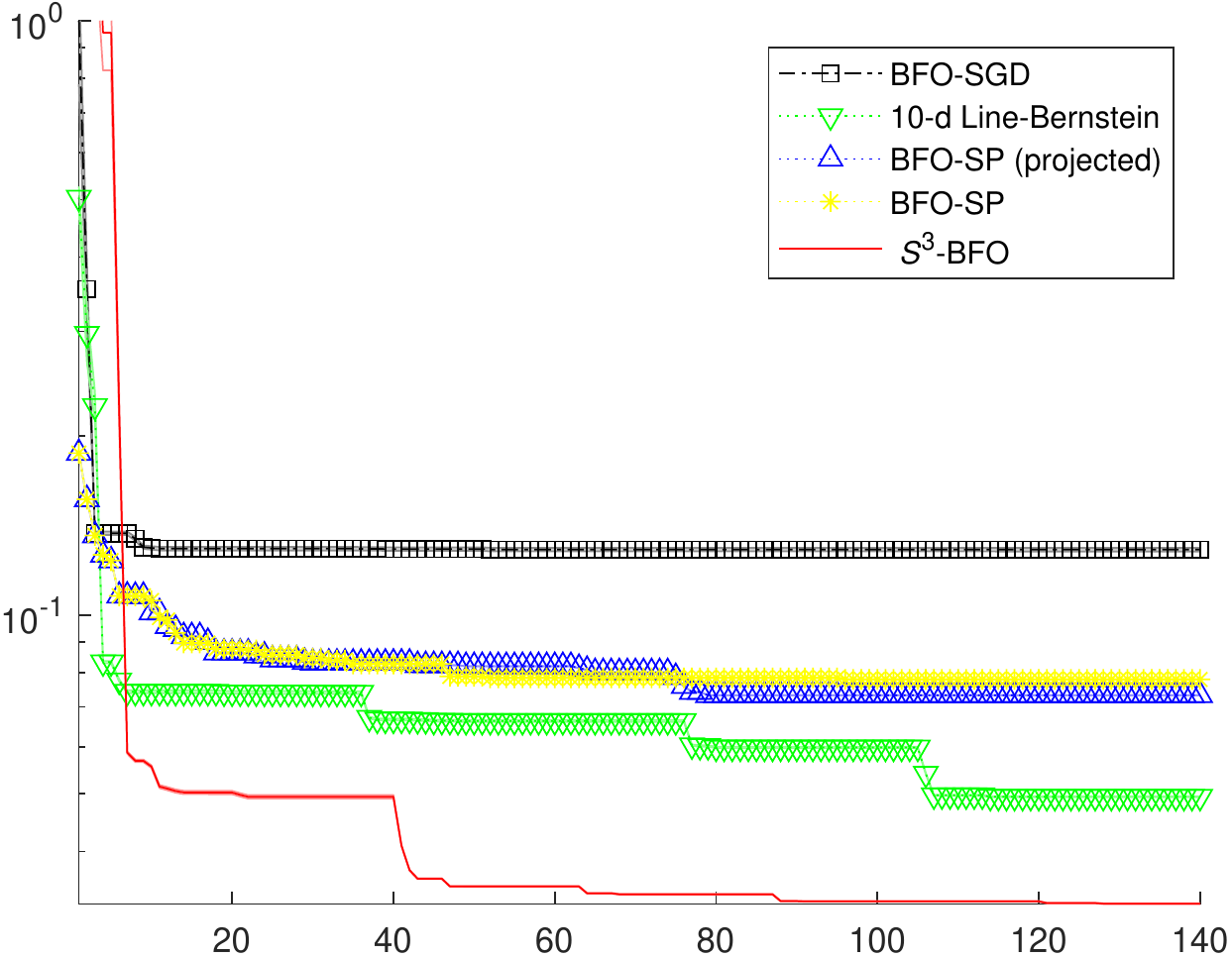}     
 $\;$ \begin{turn}{90}
 $\;\;\;\;\;\;\;\;\;\;\;\;\;$ {\small $\scriptscriptstyle{\| \func{g}-\func{q} \|_{L_2 ([0,1])}}$}
 \end{turn}
  \includegraphics[width=0.27\textwidth]{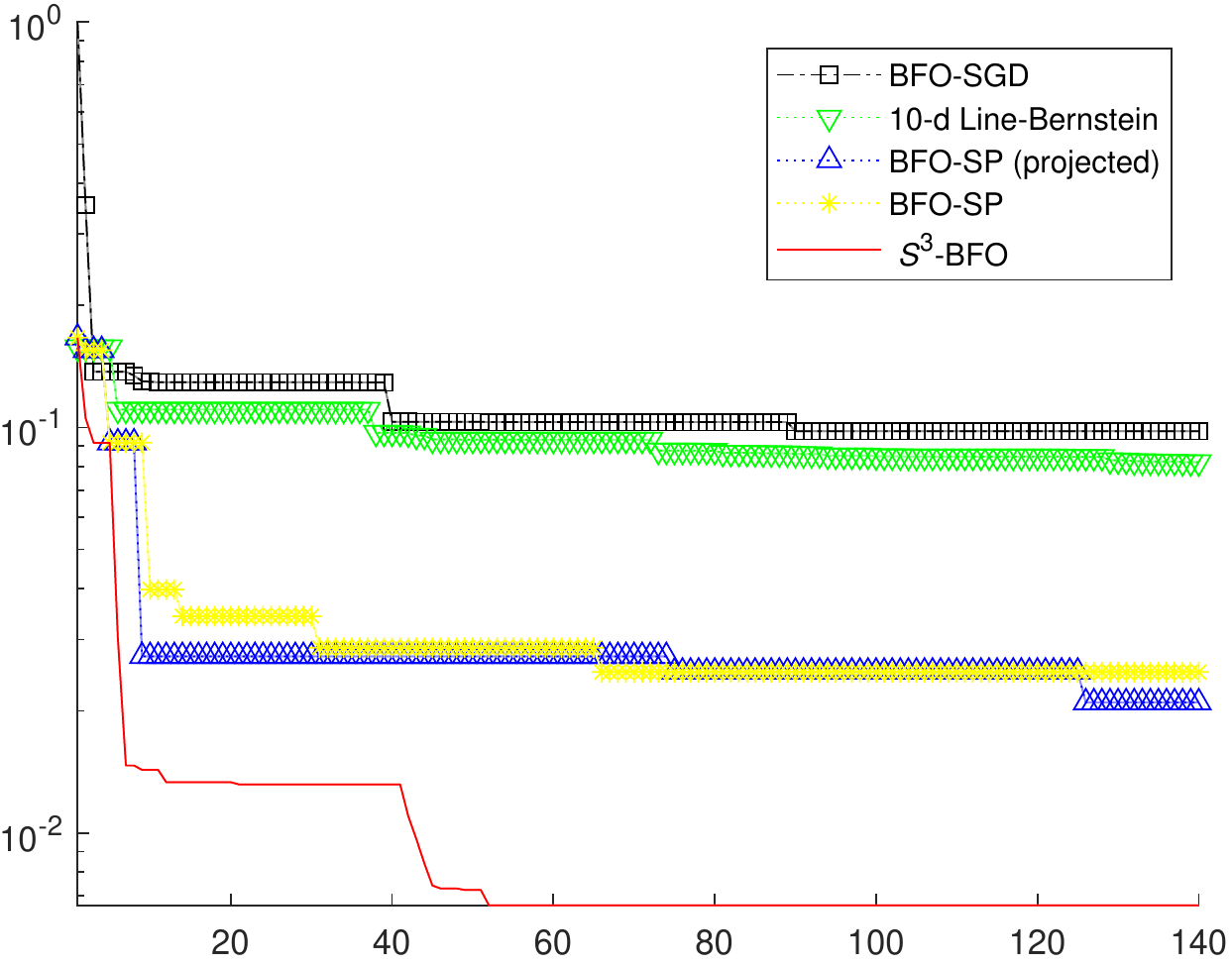}   \\
  $\;$
  \hspace{0.5cm}
  {{\small Iteration } $\scriptstyle{t+sS}$} 
  \hspace{4cm}
  {{\small Iteration } $\scriptstyle{t+sS}$} 
  \hspace{3.3cm}
  {{\small Iteration } $\scriptstyle{t+sS}$} \\
 \begin{turn}{90}
 $\;\;\;\;\;\;\;\;\;\;\;\;\;$ {\small $\scriptscriptstyle{\| \func{g}-\func{q} \|_{L_2 ([0,1])}}$}
 \end{turn}
  \includegraphics[width=0.27\textwidth]{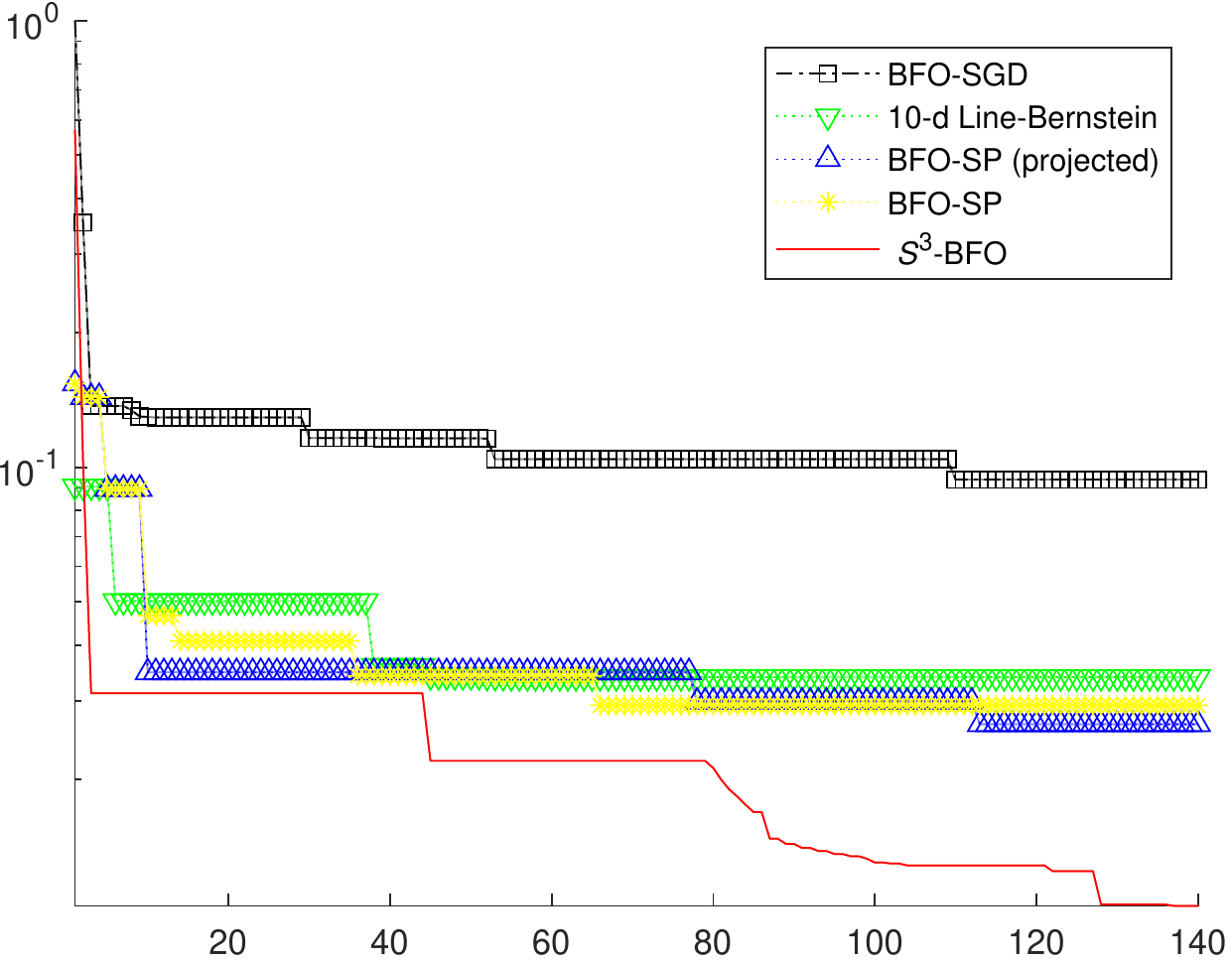}
 $\;$ \begin{turn}{90}
 $\;\;\;\;\;\;\;\;\;\;\;\;\;$ {\small $\scriptscriptstyle{\| \func{g}-\func{q} \|_{L_2 ([0,1])}}$}
 \end{turn}
  \includegraphics[width=0.27\textwidth]{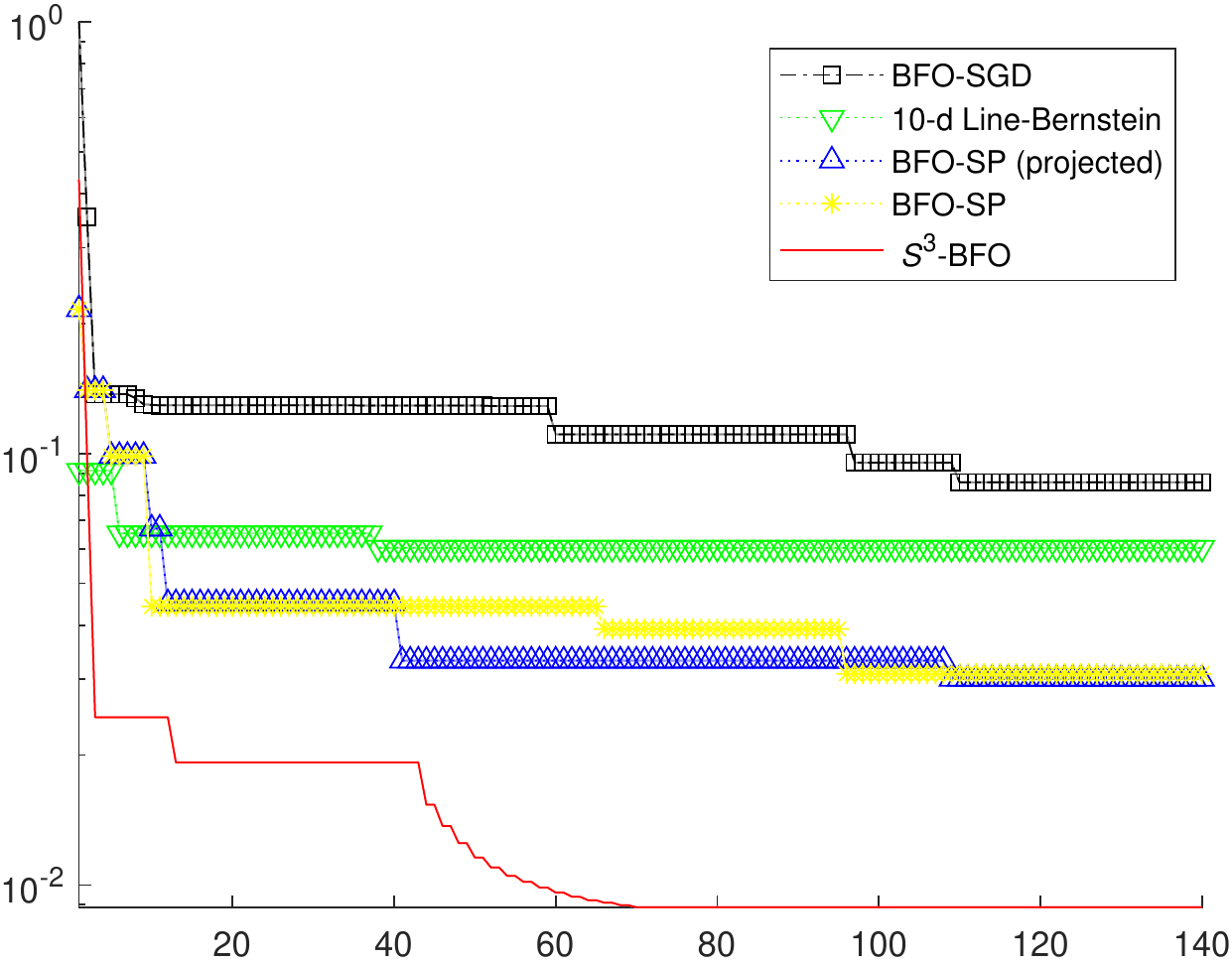} 
  $\;$ $\;\;\;$
  \includegraphics[width=0.27\textwidth]{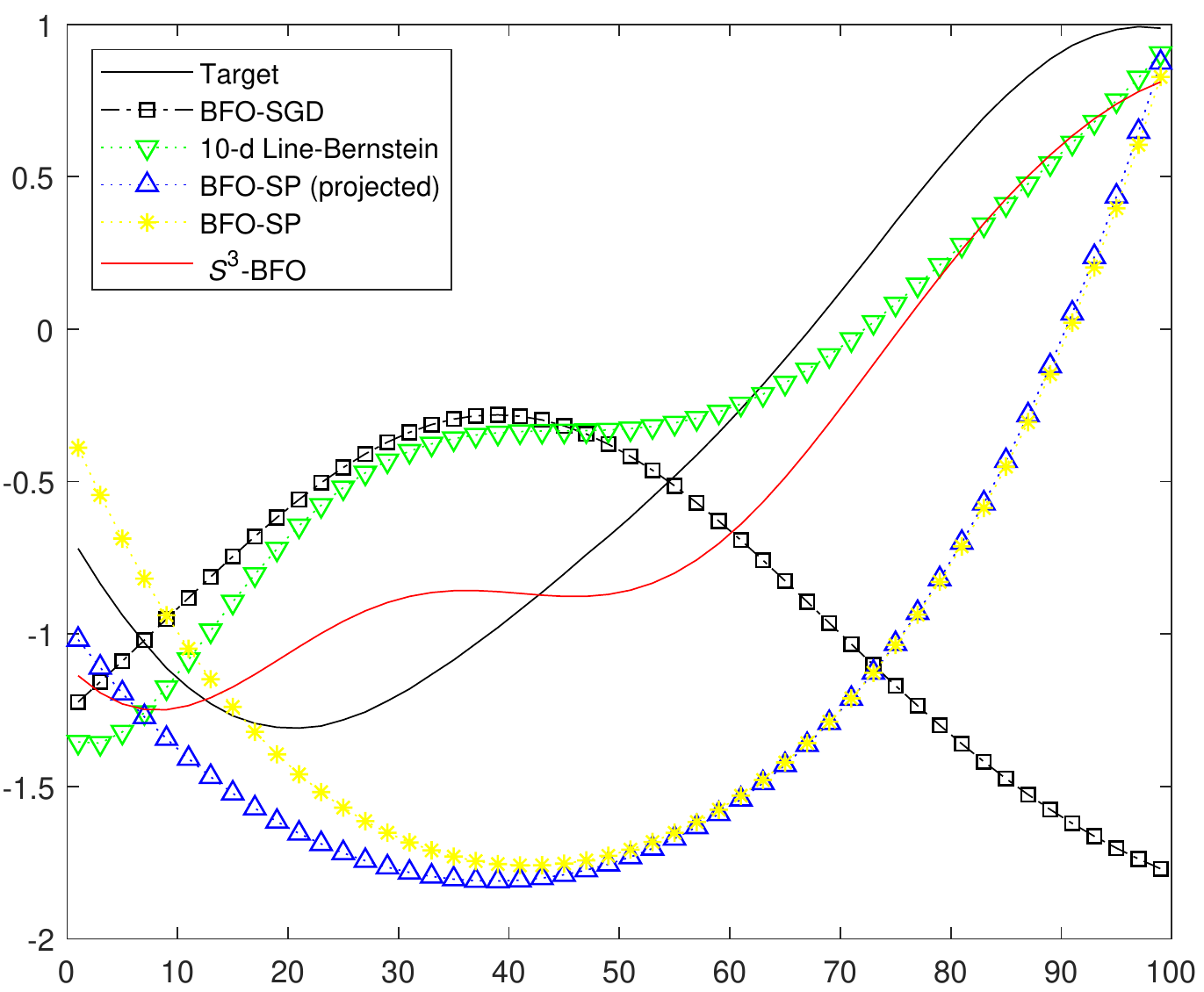} \\
  $\;$
  \hspace{0.5cm}
  {{\small Iteration } $\scriptstyle{t+sS}$} 
  \hspace{4cm}
  {{\small Iteration } $\scriptstyle{t+sS}$} 
  \hspace{5cm}
  $\;$
 \caption{Results for simulated experiments.  The top row shows the convergence 
          of $\| \func{g}-\func{q} \|_{L_2 ([0,1])}$ for three draws with the SE kernel with $\gamma = 1$, $\gamma 
          = 0.3$ and $\gamma = 0.1$ for the different algorithms, while the 
          bottom row shows same for Matern 1/2 and 3/2 (with $\gamma = 0.3$), respectively, and 
          examples of functions found by the various algorithms compared to the 
          target function for a draw from SE GP with $\gamma = 0.3$ (corresponding solutions to top row, centre).}
 \label{fig:sim1}
 \label{fig:sim1eg}
\end{figure*}

Note that, unlike LineBO, we do not build-in a requirement that the inner loop 
terminate with ${\rm err} (\func{g}_s^\star) < \epsilon$ for some fixed 
$\epsilon$.  Instead, $\epsilon_{d,\delta}$ is used, which is the (order-of) 
regret bound on the inner loop.  If the simple regret test is implemented on 
the inner loop then $\epsilon_{d,\delta} = \epsilon$.  An alternative, fixed 
budget strategy is also discussed below.  
Also unlike LineBO the exponent in the regret bound includes the dimension $d$ 
of the random subspaces.  At one extreme, if $d = 1$ and we use 
the inner-loop convergence condition ${\rm err} (\func{g}_s^\star) < \epsilon$, 
then theorem \ref{thm:th_converge} is essentially the same as Proposition 1 in 
\cite{Kir2} for LineBO with $\epsilon_{d,\delta} = \epsilon$.  In this case the 
inner loop may be expected to take $T \in \bigo 
(\epsilon^{\frac{2}{1-2\kappa}})$ iterations to complete, and the overall 
number of function evaluations required by the algorithm is $\bigo (S 
\epsilon^{\frac{2}{1-2\kappa}})$ \cite{Kir2}, where $\kappa \in (0, 0.5)$ is a 
function of the covariance $K$.  If instead $d = 1$ and the inner loop is 
allocated a fixed budget of $T$ iterations then we find $\epsilon_{1,\delta} 
\in \bigo (T^{\kappa-\frac{1}{2}})$, and the algorithm will make precisely 
$ST$ function evaluations.  
At the opposite extreme, if $d = d_e$ then the first term in the regret bound 
in theorem \ref{thm:th_converge} disappears entirely and the regret is entirely 
due to $\epsilon_{d_e,\sigma}$.  This case is analogous to REMBO, where we know 
that, with probability $1$, any random basis suffices to capture the necessary 
variation of $f$.  Note that in this case we may set $S = 1$ without affecting 
our regret bound.  The regret bound in this case collapses to precisely the 
standard regret bounds found in for example \cite{Sri1}.

Between these extremes the algorithm may be expected to act somewhat like a 
combination of LineBO and REMBO, although of course if $d_e$ is too large - say 
$d_e \gtrsim 10$ - then setting $d = d_e$ will not be practical, so in this 
case multiple outer-loop iterations ($S > 1$) will be required.

\section{Experimental Results}

We consider simulated and real-world experiments.  In our simulated experiment 
we take a draw $\func{q}$ from a GP and then attempt to reconstruct this draw 
(that is, find $\func{g}$ such that $\func{g} = \func{q}$) without explicit 
knowledge of $\func{q}$, but with the ability to test/calculate $\| \func{g} - 
\func{q} \|_{L_2(\infset{A})}$.
  Our two real-world experiments are finding the optimal precipitation 
strengthening function for a metallic alloy of Aluminium and finding the 
optimal rate scheduling for deep network training.

All optimisers were implemented in and run with SVMHeavy v7 \cite{svmheavyv7} 
(code available at \url{https://github.com/apshsh/SVMHeavy}), excepting the 
KWN implementation, which is proprietary at present.

We have compared our method with the following: 10-d-Line-Bernstein, which 
tunes the weights of a $10^{\rm th}$ order Bernstein polynomial approximation 
using LineBO; BFO-SP, which implements Vellanki's algorithm \cite{Vel1}; BFO-SP 
projected, which is like BFO-SP but models $f$ in function space as per our 
algorithm; and BFO-SGD, which is our implementation of \cite{Vie1}'s algorithm.  
All models were based on variants of the SE kernel (real, $L_2$ or RKHS), and 
all experiments were repeated $5$ times to generate error bars.  See 
supplementary for further details on experimental setup.

\subsection{Simulated experiment}

The aim of this experiment is to investigate 
the role of experimenter beliefs in functional optimisation, so for the 
purposes of this experiment we assume that the experimenter has a good 
intuitive understanding of the expected lengthscale, smoothness etc (in the 
form of a covariance function) that cannot be directly used in the alternative 
methods.  To achieve this, we consider function reconstruction - that is, given 
a target function $\func{q} : \left[ 0,1 \right] \to \infset{R}$, we aim to 
solve the functional optimisation problem:
\[
 \begin{array}{l}
  \func{g}^* = {\rm argmax}_{\func{g} : {[0,1]} \to \infset{R}} \left\| \func{g}-\func{q} \right\|_{L_2 ([0,1])}
 \end{array}
\]

As our target functions we have used draws from three Gaussian process, 
$\func{q} \sim \gp (0,k)$, where $k (x,x') = \exp (\frac{1}{2\gamma} (x-x')^2)$ 
and $\gamma = 1$, $\gamma = 0.3$ and $\gamma = 0.1$, respectively, as well a 
draw from a Gaussian process $\func{q} \sim \gp (0,\kappa)$, where $\kappa$ is 
Matern-$\nicefrac{1}{2}$ kernel (with $\gamma = 0.3$), and another where 
$\kappa$ is a Matern-$\nicefrac{3}{2}$ kernel (with $\gamma = 0.3$).   In these 
experiments we assume the experimenter has a good intuition regarding the 
target function, so the covariance $\kappa$ is the same as the GP $k$ from 
which the target was drawn.

Figure \ref{fig:sim1} shows convergence results for the functions drawn from 
a GP for different lengthscales, along with a sample of the functions found by 
the difference approaches (first run in sequence).  We note that most methods 
perform reasonably for the longest lengthscale, which represents the simplest 
function to approximate.  As the lengthscale is shortened we see that our 
method continues to perform well due to the incorporation of experimenter 
knowledge, while the alternatives become progressively less accurate.  For 
10-d-Line-Bernstein this is because the Bernstein polynomial of that order is 
unable to capture the complexity resulting from the shorter length-scale; and 
likewise for the BFO-SP variants, although the algorithm is designed to tune 
the complexity as required, this takes some time, whereas in our algorithm the 
experimenter intuition is built in.
Finally, figure \ref{fig:sim1}, bottom right, shows the best functions found by 
each algorithm in the first simulation run.

\subsection{Precipitation Strengthening in Al-Sc Alloy}

Heat treatment of alloys makes them stronger by providing a desired grain 
structure through precipitation of different crystal structures. Normally, the 
alloy is heated to a high temperature to first homogenise the structure, then 
taken through a series of temperature to achieve desired pattern of 
precipitates.  For Al-Sc alloy precipitation strengthening has been proven to 
be particularly effective \cite{Kni1,Kni2,Sei1}; however, as scandium is 
expensive and the experimental process time-consuming there has been relatively 
little work in the determination of optimal temperature profile 
\cite{Dea2,Vah1,Vel2}.

\begin{figure}
 \centering 
 \begin{turn}{90}
 $\;\;\;\;\;\;\;\;\;\;\;\;\;\;\;$ {\tiny Alloy strength}
 \end{turn}
  \includegraphics[width=0.3\textwidth]{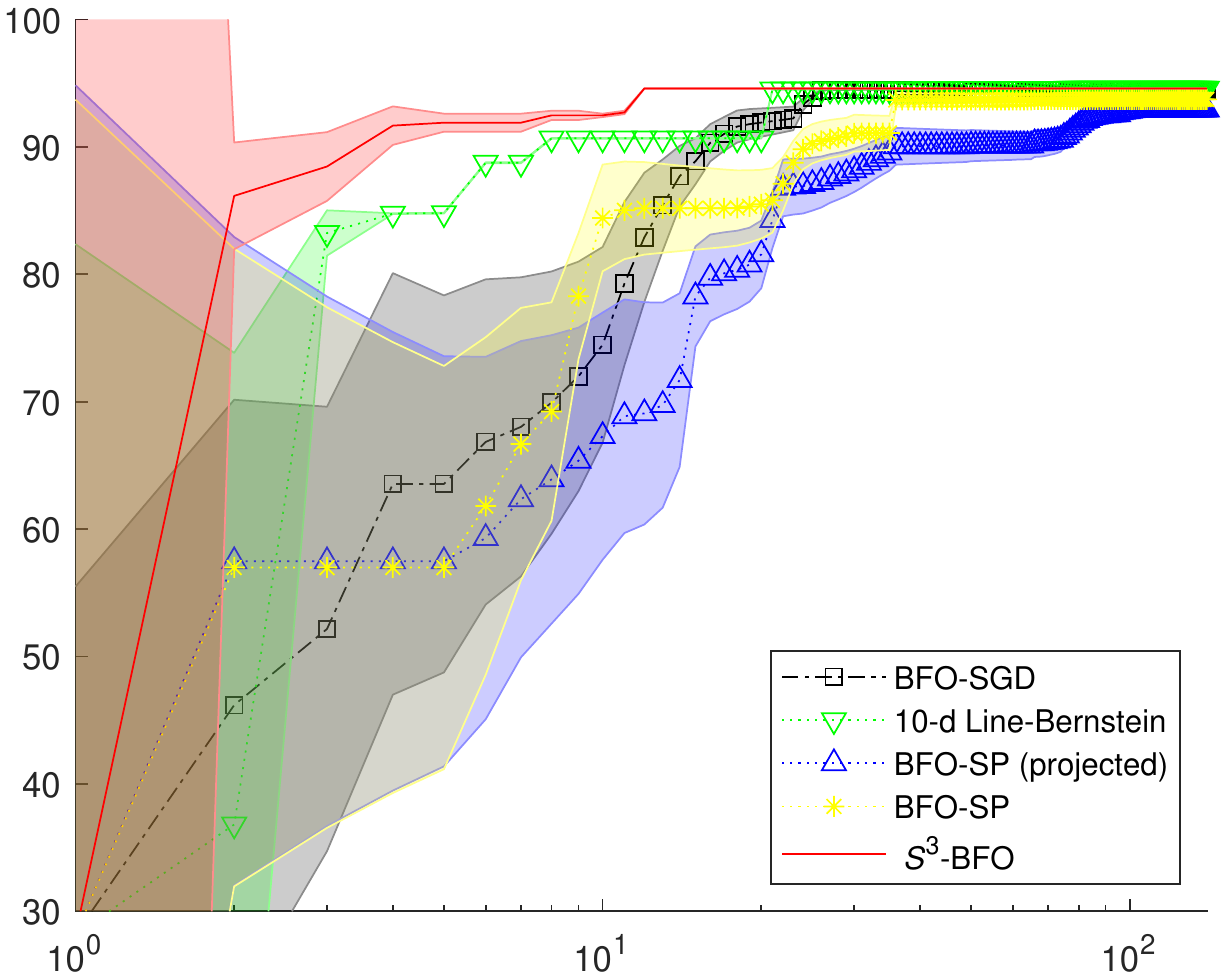} \\
  {{\small Iteration } $\scriptstyle{t+sS}$} \\
 \caption{Convergence of precipitation-strengthening.  As the alloy strength is 
          highly dependent on the temperature profile there is large variance 
          in strength in the early stages depending on the (random) set of 
          initial observations.}
 \label{fig:kwnsim}
\end{figure}

In this experiment we model the precipitation-strengthening process using 
Kampmann and Wagner's numerical model (KWN) \cite{Wag3,Kni3}.  This model was 
implemented in MATLAB and allows us to predict the final strength of the alloy 
processed according to a given temperature profile (suitably discretized, in 
our experiments using $100$ timepoints).  As shown in figure \ref{fig:kwnsim}, 
our algorithm converges more quickly than the alternatives.  As for the 
simulated experiment, in this case BFO-LB$10$ (LineBO to tune the weights of a 
$10^{\rm th}$ order Bernstein polynomial) converges second fastest, followed 
by Vellanki's method (BFO-SP and BFO-SP (projected)).

\subsection{Learning Rate Schedule Optimisation}

As noted in \cite{Ben11}, and following \cite{Vel1}, stochastic gradient 
descent (SGD) works better if the learning rate is varied as a function of 
training duration.  In this experiment we optimise the learning rate for a 
neural network trained on the MNIST dataset (other parameters being kept 
constant).  As a baseline we compare results achieved with our method, and the 
other baselines already described, with SGD using learning rate $0.1$ (decaying 
exponentially as per \cite{Vel1}) and momentum $0.8$, and Adam with default 
hyper-parameters \cite{Kin3}.  Unlike \cite{Vel1} we do not enforce a 
decreasing rate prior for any of the methods compared.

\begin{figure}
\centering
 \includegraphics[width=0.29\textwidth]{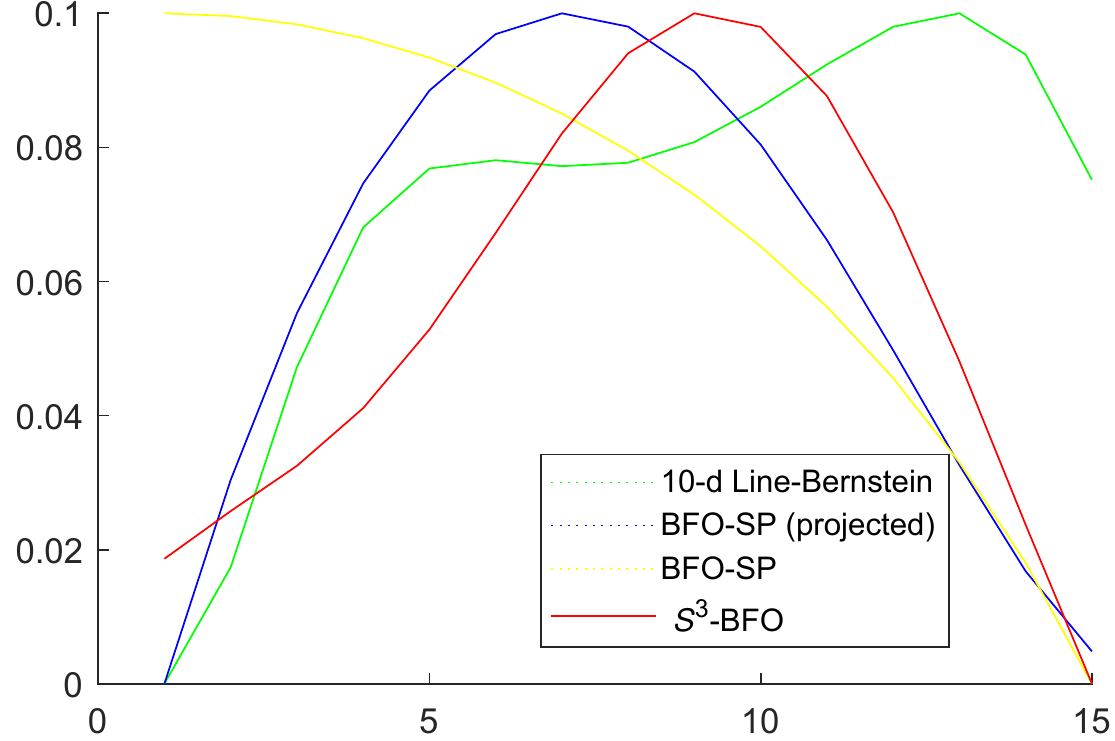} \\
 \caption{Scheduling functions found by optimisers.}
 \label{fig:scheduleres}
\end{figure}

Results are shown in table \ref{tab:lrschedres} (results for SGD, Adam and 
BFO-SGD are sourced from \cite{Vie1,Vel1}).  Note that our method achieves the 
lowest validation error of all approaches considered (\cite{Vel1} achieves a 
better result, but only be applying decreasing prior).  Actual learning rate 
schedules are shown in figure \ref{fig:scheduleres} (learning rate schedule for 
BFO-SGD can be found in \cite{Vel1,Vie1}.  It is perhaps interesting to note 
that most solutions found are a simple rise/fall function with a single peak.

\begin{table}
\centering
\begin{tabular}{| l | l |}
\hline
SGD                 & $1.26\%$ \\
\hline
Adam                & $0.86\%$ \\
\hline
BFO-SGD${}^*$       & $0.87\%$ \\
\hline
10-d Line-Bernstein & $0.78\%$ \\
\hline
BFO-SP (projected)  & $0.78\%$ \\
\hline
BFO-SP              & $0.77\%$ \\
\hline
$S^3$-BFO           & ${\bf 0.76\%}$ \\
\hline
\end{tabular}
\caption{Validation error of MNIST neural network for different learning rate 
         schedules (${}^*$results from \cite{Vie1}).}
\label{tab:lrschedres}
\end{table}

\section{Conclusion}

We have proposed an algorithm for Bayesian functional optimisation - that is, 
optimisation problems such as finding the best temperature profile for alloy 
heat treatment, where experiments are expensive/time-consuming and results are 
noisy.  Our algorithm allows the experimenter to express prior beliefs regarding 
the solution in the form of a covariance function, specifying e.g. length-scale, 
smoothness, etc.  Guided by this prior, our algorithm generates a sequence of 
finite-dimensional random subspaces, applying standard Bayesian optimisation on 
each and then building the next subspace from the best solution found.  We have 
presented a sub-linear regret bound for our algorithm and provided experimental 
results on simulated and real-world experiments, namely simulated function 
mapping, finding the optimal precipitation-strengthening function for an 
aluminium alloy, and learning-rate scheduling for deep-network training.

\newpage
{\small
\bibliographystyle{plainnat}
\bibliography{universal}
}

\newpage
\section{Supplementary: Caching and Optimisation} \label{sec:sampdetail}

To improve calculation speed, in our implementation we pre-define a grid $\{ 
{\bf c}^i \in \infset{A} = [0,1]^m \subset \infset{R}^m | i \in \infset{N}_{N} \}$ on 
$\infset{A}$ as an even grid of $N^{1/m}$ points per axis in $\infset{A} 
\subset \infset{R}^m$.  Specifically, assuming 
$\infset{A}$ is an $m$-dimensional unit hypercube, an even grid of $N^{1/m}$ points 
per axis with spacing $\tau = N^{-1/m}$ (see below).  This allows us to approximate points 
$\func{g} \in L_2 (\infset{A})$ using ${\bf g} \in \infset{R}^N$.

The set of initial observations $\finset{D}_{[0]}$ are sampled on our grid 
before entering the algorithm, so $\func{g}$ is represented as ${\bf g}$, 
$g_i = \func{g} ({\bf c}_i)$, $\forall (\func{g},y) \in \finset{D}_{[0]}$, 
which also gives us a sampled form ${\bf b}_0 = {\bf g}_{[0]}^\star$ of 
$\func{b}_g$ for the first (outer) iteration ($s = 0$).  To draw basis 
functions $\func{h}_s^{j} \sim \gp (0,\kappa)$ in our algorithm we 
draw an $N$-dimensional vector:
\[
 \begin{array}{l}
  {\bf h}_s^{j} := 
  \left[ \begin{array}{c} 
  \func{h}_s^{j} \left( {\bf c}^0 \right) \\ 
  \func{h}_s^{j} \left( {\bf c}^1 \right) \\ 
  \vdots \\ 
  \end{array} \right] 
  \sim \normdist \left( {\bf 0},
  \left[ \begin{array}{ccc} 
  \kappa \left( {\bf c}^0,{\bf c}^0 \right) & \kappa \left( {\bf c}^0,{\bf c}^1 \right) & \cdots \\
  \kappa \left( {\bf c}^1,{\bf c}^0 \right) & \kappa \left( {\bf c}^1,{\bf c}^1 \right) & \cdots \\
  \vdots & \vdots & \ddots \\
  \end{array} \right] \right)
 \end{array}
\]
This allows us to evaluate $\func{g}$ on our grid (and cache for later):
\[
 \begin{array}{l}
  {\bf g} := \left[ \; \func{g} \left( {\bf c}^0 \right) \; \func{g} \left( {\bf c}^1 \right) \; \ldots \; \right]^{\tsp} = {\bf b}_s + \sum_j \lambda^t_j {\bf h}^{j}
 \end{array}
\]
and the process may be repeated for subsequent iterations.  It follows that we 
can easily approximate:
\begin{equation}
 \begin{array}{l}
  \| \func{g}-\func{g}' \|_{L^2 (\infset{A})}^2 \approx \| {\bf g} - {\bf g}' \|_2^2 \tau^m
 \end{array}
 \label{eq:approxL2norm}
\end{equation}
and thus avoid the need to (a) repeatedly re-evaluate our basis functions 
$\func{h}_s^{j}$ on our grid and (b) calculate lengthy weighted sums of basis 
elements when evaluating $\func{g}$ or evaluating the covariance matrix on 
our grid.

\vspace{-0.1cm}
The number of points $N$ required to attain reasonable accuracy in our 
approximation (\ref{eq:approxL2norm}) depend on the characteristics of the 
(tested) functions $(\func{g},\cdot) \in \finset{D}$.  In our algorithm these 
lie in the span of a set of draws from $\gp (0,\kappa)$, and so inherit their 
characteristics from the covariance prior $\kappa$.  Thus if $\kappa$ is has 
length-scale $\gamma$ then it seems reasonable to select $N \approx 
(D/\gamma)^m$ for some constant $D$.  However this does not {\em guarantee} the 
accuracy of (\ref{eq:approxL2norm}), as $\func{g} \sim \gp (0,\kappa)$ only 
implies that $\func{g}$ is likely to have characteristics suitable for 
such an approximation - for example, for an SE kernel $\kappa$, the space 
of possible draws from $\gp (0,\kappa)$ is {\em independent} 
of the length-scale $\gamma$ (the SE kernel is universal \cite{Mic1,Sri2}), so 
even if the particular prior used in our algorithm has a long length-scale 
$\gamma$ it is possible (though unlikely) that our algorithm will explore 
regions that vary on a much shorter scale than we might naively expect.  In 
practice we recommend being generous when selecting $N$ as the complexity of 
all relevant operations in our algorithm (weighted sums and 
(\ref{eq:approxL2norm})) scale linearly with $N$, so the penalty for 
``overdoing it'' to ensure accurate approximation in (\ref{eq:approxL2norm}) is 
relatively small (for example in our experiments we use $N = 100$).

\section{Supplementary: Convergence Analysis - Proof of Theorem 1}

We begin by proving some preliminary results.  We define $\sgp{d} (\func{b})$ 
to be the distribution of random subspaces of $L^2 (\infset{A})$ of the form 
$\func{b} + \spn (\func{h}^0, \func{h}^1, \ldots, \func{h}^{d-1})$, where 
$\func{b}$ is some fixed ``origin'' point and $\func{h}^0, \func{h}^1, \ldots, 
\func{h}^{d-1} \sim \gp (0,\kappa)$.  For each outer-loop iteration $s$ of 
algorithm 2, the inner loop performs Bayesian Optimisation on a subspace 
$\infset{U}_s \sim \sgp{d} (\func{b}_s)$.  For notational convenience we define 
$\infset{U}_{[s]} = \cup_{i \in \infset{N}_s} \;\infset{U}_{i}$, ${\bf x}_{s}^* 
= \arg\min_{{\bf x} \in \infset{U}_{s}} f ({\bf x})$, and ${\bf x}_{[s]}^* = 
\arg\min_{{\bf x} \in \infset{U}_{[s]}} f ({\bf x})$.  We have the results (it 
seems probable that lemma \ref{lem:lem_geom_arg} is ``well known'', but we have 
been unable to find a reference):
\begin{lem_geom_arg}
 Let $\infset{U} \subseteq \infset{V} = \{ {\bf v} \in \infset{R}^{d_e} | \| 
 {\bf v} \|_2 \leq L \}$, where $\infset{U} = \spn ({\bf u}^0, {\bf u}^1, 
 \ldots, {\bf u}^{d-1}) + {\bf b} \cap \infset{V}$, ${\bf u}^i \perp {\bf u}^j$ 
 $\forall i \ne j \in \infset{N}_d$, ${\bf u}^i \sim \distrib{U}_i$ $\forall i 
 \in \infset{N}_d$, ${\bf b} \sim \distrib{B}$, for smooth distributions 
 $\distrib{U}_i, \distrib{B}$.  Then the probability that $\infset{U}$ 
 intersects the $d_e$-ball of radius $r = \beta L$, $\beta \in (0,1]$, at the 
 origin is at least $\Omega (\beta^{d_e-d})$ if $d < d_e$, $1$ otherwise.
 \label{lem:lem_geom_arg}
\end{lem_geom_arg}
\begin{subproof}
 Denote the probability of intersection by $\zeta_{d,d_e} (\beta)$.  With 
 probability $1$ we have that $\| {\bf u}^i \|_2 \ne 0$ $\forall i \in 
 \infset{N}_{d}$.  Hence we may assume ${\rm dim} (\infset{U}) = d$.
 
 If $d = d_e$ then $\infset{U} = \infset{V}$ and $\zeta_{d_e,d_e} (\beta) = 1$ 
 trivially.  If $d = 0$ then the probability of intersection is precisely the 
 probability that a point selected from a smooth distribution falls into an 
 $d_e$-ball of radius $r = \beta L$, which goes as the ratio of the measure of 
 the $d_e$-ball and the measure of $\infset{V}$ - that is, $\zeta_{0,d_e} (\beta) 
 = \Omega (\beta^{d_e})$.

 Otherwise if $0 < d < d_e$ note that, as both $\infset{V}$ and the $d_e$-ball 
 are rotationally symmetric about the origin, we may always assume that ${\bf 
 u}^{i} = [\; \delta_{0,i} \; \delta_{1,i} \; \ldots \; \delta_{d-1,i} \; {\bf 
 0} \;] \tilde{u}_i$, $i \in \infset{N}_d$, where $\delta_{i,j}$ is the 
 Kronecker-delta symbol.  Hence, writing ${\bf b} = [\; \hat{\bf b} \; 
 \check{\bf b} \;]$, $\hat{\bf b} \in \infset{R}^d$, $\check{\bf b} \in 
 \infset{R}^{d_e-d}$, and noting that $\tilde{u}_i \ne 0$ $\forall i \in 
 \infset{N}_d$ with probability $1$ and $\check{\bf b} \sim 
 \check{\distrib{B}}$ (conditioned on $\hat{\bf b}$), we have with probability 
 $1$:
 \[
  \begin{array}{rl}
   \zeta_{d,d_e} (\beta) 
   &\!\!\!= \Pr \left( \mathop{\min}\limits_{{\mbox{\boldmath $\gamma$}} \in \infset{R}^d} \left\| \left[ \begin{array}{c} \hat{\bf b} + {\mbox{\boldmath $\gamma$}} \odot \tilde{\bf u} \\ \check{\bf b} \\ \end{array} \right] \right\|_2 \leq \beta L \right) \\
   &\!\!\!= \Pr \left( \left\| \check{\bf b} \right\|_2 \leq \beta L \right) \\
  \end{array}
 \]
 where $\odot$ is the elementwise product and the minima is attained with 
 $\gamma_i = -\hat{b}_i/\tilde{u}_i$ $\forall i \in \infset{N}_d$.  However 
 this is precisely equivalent to the $d = 0$ case with decreased $d_e$, hence 
 $\zeta_{d,d_e} (\beta) = \zeta_{0,d_e-d} (\beta) = \Omega(\beta^{d_e-d})$ when 
 $0 < d < d_e$.
\end{subproof}

\begin{lem_analog_kir2_lemma2}[analog of \cite{Kir2}, Lemma 2] 
For outer-loop iteration $s$ of algorithm 2:
 \[
  \begin{array}{l}
   \Pr \left( f \left( \func{g}^* \right) - f \left( \func{g}_{[s]}^* \right) \leq \tau \right) \geq 1 - \exp \left( -s \xi \left( \tau \right) \right)
  \end{array}
 \]
 where $\xi (\tau)$ is a lower bound on:
 \[
  \begin{array}{l}
   \xi \left( \tau \right) \leq \Pr \left( \left. \exists \func{g} \in \infset{U}, f \left( \func{g}^* \right) - f \left( \func{g} \right) \leq \tau \right| \infset{U} \sim \sgp{d} \left( \func{b} \right) \right)
  \end{array}
 \]
 Furthermore if the first-order minimum condition is met at $\func{g}^*$ then 
 $\xi (\tau) = \Omega ( \tau^{\frac{d_e-d}{2}})$ if $d < d_e$, $\xi (\tau) = 
 1$ otherwise.
\end{lem_analog_kir2_lemma2}
\begin{subproof}
 The proof follows the approach of \cite{Kir2}, extended to the functional 
 domain. Using the inequality $1-x \leq e^{-x}$, we have that:
 \[
  \begin{array}{rl}
   \Pr ( f ( \func{g}^* ) - f ( \func{g}_{[s]}^* ) \leq \tau ) 
   &\!\!\!= 1 - \Pr ( f ( \func{g}^* ) - f ( \func{g}_{[s]}^* ) \geq \tau ) \\
   &\!\!\!= 1 - {\prod}_{i\in\infset{N}_s} \Pr \left( f \left( \func{g}^* \right) - f \left( \func{g}_{i}^* \right) \geq \tau \right) \\
   &\!\!\!\geq 1 - \left( 1 - \xi \left( \tau \right) \right)^s \\
   &\!\!\!\geq 1 - \exp \left( -s \xi \left( \tau \right) \right) \\
  \end{array}
 \]

 Recall that, by definition, $\exists \bar{\func{h}}^0, \bar{\func{h}}^1, 
 \ldots, \bar{\func{h}}^{d_e-1} \in L^2(\infset{A})$ such that $\| f 
 (\func{g}_{\top} + \func{g}_{\perp}) - f (\func{g}_{\top}) \|_{L^2 
 (\infset{A})} = 0$ $\forall \func{g}_{\top} \in \infset{T}$, $\forall 
 \func{g}_{\perp} \in \infset{T}^{\perp}$, where $\infset{T} = \spn 
 (\bar{\func{h}}^0, \bar{\func{h}}^1, \ldots, \bar{\func{h}}^{d_e-1})$.  
 We adopt the notational convention that $\forall \func{g} \in L^2 
 (\infset{A})$, $\func{g} = \func{g}_{\top} + \func{g}_{\perp}$ where 
 $\func{g}_{\top} \in \infset{T}$, $\func{g}_{\perp} \in 
 \infset{T}^{\perp}$.  Define:
 \[
  \begin{array}{rl}
   \infset{V}_{\tau} 
   &\!\!\!= \left\{ \left. \func{g} \in L^2 (\infset{A}) \right| f \left( \func{g}^* \right) - f \left( \func{g} \right) \leq \tau \right\} \\
  \end{array}
 \]
 to be the set of solutions within $\tau \geq 0$ of the optima.  As $f 
 (\func{g}) = f (\func{g}_\top)$ it follows that $\infset{V}_{\tau} = 
 \infset{V}_{\tau\top} \oplus \infset{T}^{\perp}$, where:
 \[
  \begin{array}{rl}
   \infset{V}_{\tau\top} 
   &\!\!\!= \left\{ \left. \func{g}_\top \in \infset{T} \right| f \left( \func{g}^* \right) - f \left( \func{g}_\top \right) \leq \tau \right\} \\
  \end{array}
 \]
 has dimension $d_e$.  Hence to place a lower bound on $\xi (\tau)$ it suffices 
 to  bound the probability that a random $d$-dimensional linear subspace 
 $\infset{U} \sim \sgp{d} (\func{b})$ projected onto $\infset{T}$ (ie. 
 $\infset{U}_{\top} = \{ \func{g}_\top | \func{g} \in \infset{U} \}$, 
 $\infset{U} \sim \sgp{d} (\func{b})$) intersects $\infset{V}_{\tau\top}$.  To 
 bound this, define:
 \[
  \begin{array}{rl}
   \tilde{\infset{V}}_{\tau,\alpha\top} 
   &\!\!\!= \left\{ \left. \func{g}_\top \in \infset{T} \right| \frac{\alpha}{2L_{\rm max}^2} \left\| \func{g}^* - \func{g}_{\top} \right\|_{L^2 \left( \infset{A} \right)}^2 \leq \tau \right\}
  \end{array}
 \]
 where $\alpha > 0$.  Using the fact that $f$ is twice Frechet differentiable 
 we have that $f (\func{g}^* + \func{q}) \geq f (\func{g}^*) - \frac{\alpha}{2L_{\rm max}^2} 
 \| \func{q} \|_{L^2 (\infset{A})}^2$ for sufficiently small $\frac{\alpha}{2L_{\rm max}^2} \| 
 \func{q} \|_{L^2 (\infset{A})}^2$.  Letting $\func{q} = \func{g}_\top - 
 \func{g}^*$ we see that $f (\func{g}^*) - f (\func{g}_\top) \leq 
 \frac{\alpha}{2L_{\rm max}^2} \| \func{g}_\top - \func{g}^* \|_{L^2 (\infset{A})}^2$, so 
 $\tilde{\infset{V}}_{\tau,\alpha\top} \subseteq \infset{V}_{\tau\top}$.
 
 Hence to place a lower bound on $\xi (\tau)$ it suffices to bound the 
 probability that a random $d$-dimensional linear subspace $\infset{U}_\top$ in 
 a $d_e$-dimensional space intersects a $d_e$-ball at the origin of radius 
 $\sqrt{2\tau/\alpha}L_{\rm max}$, which by Lemma \ref{lem:lem_geom_arg} is 
 $\Omega (\tau^{\frac{d_e-d}{2}})$ if $d < d_e$ and $1$ otherwise, 
 completing the proof.
\end{subproof}

Having established the above result the proof of theorem 1 follows almost 
precisely that of \cite{Kir2}, proof of proposition 1, excepting that $d_e-1$ 
is replaced by $d_e-d$ wherever present, and rather than enforcing an upper 
bound on $\epsilon$ we allow it to vary with order $\epsilon_{d,\delta}$.

\section{Supplementary: Details of Experimental Procedure}

In the paper we have compared the following methods:
\begin{enumerate}
  \item $S^3$-BFO: our method as described, using a $d = 1$ dimensional 
       search subspace, $5$ initial observations, $S = 4$ outer loop 
       iterations, $T = 30$ inner loop iterations (fixed budget on the inner 
       loop), $L^2 (\infset{R})$-SE covariance prior $K$ on $f$ 
       with length-scale selected for maximum 
       likelihood at each model update, SE covariance prior $\kappa$ on 
       $\func{g}^*$ with length-scale $0.3$ unless otherwise stated, GP-UCB 
       acquisition function.
 \item $10$-d-Line-Bernstein: LineBO algorithm used to tune the weights of $10^{\rm th}$-order Bernstein 
       polynomial, using $4$ lines in sequence, $30$ iterations per 
       line (standard GP-UCB BO on each line), where $f$ is modelled on weights of Bernstein 
       polynomial using SE covariance $K$ with length-scale selected for 
       maximum likelihood at each iteration.
 \item BFO-SP: Vellanki's algorithm \cite{Vel1}, with $f$ modelled on weights of 
       Bernstein polynomial using SE covariance $K$ with length-scale selected 
       for maximum likelihood at each iterations.
 \item BFO-SP (projected): like BFO-SP, except that in this case $f$ has been 
       modelled in function space using $L^2 (\infset{R})$-SE covariance prior 
       with length-scale selected for maximum 
       likelihood at each iteration.
 \item BFO-SGD: based on \cite{Vie1}, $f$ modelled in function space 
       using $\rkhs{\kappa}{\infset{R}}$-SE covariance prior 
       (as per \cite{Vie1}) with length-scale selected 
       for maximum likelihood at each iteration, $\kappa$ is SE covariance 
       with length-scale $0.3$ unless otherwise stated.  Note that, while we were 
       unable to obtain source code from the authors, every effort has been made 
       to ensure that our implementation matches the description as closely as 
       possible.
\end{enumerate}

When calculating norms in $L^2$ we have used a uniform grid with spacing 
$\tau = 0.01$ (so $N = 100$ for a $1$-dimensional function) - see section 
\ref{sec:sampdetail} for more information on how this effects our simulation.  
All experiments were repeated $5$ times to obtain error bars.  All optimisers 
and simulators were implemented in and run with SVMHeavy v7 \cite{svmheavyv7} 
and a proprietary KWN implementation.  SVMHeavy is available on github at 
\url{https://github.com/apshsh/SVMHeavy}.

\end{document}